\documentclass{article}

\usepackage[misc]{ifsym}
\usepackage{microtype}
\usepackage{graphicx}
\usepackage{subfigure}
\usepackage{booktabs} 
\usepackage{soul}
\usepackage{float}

\usepackage{hyperref}

\usepackage{adjustbox}
\usepackage{multirow}

\usepackage{pifont}

\usepackage[accepted]{icml2021}
\icmltitlerunning{Lottery Ticket Preserves Weight Correlation: Is it Desirable or Not?}

\begin{document}

\newcommand{\GY}[1]{\textcolor{blue}{Geng: #1}}

\newcommand{\todo}[1]{\textcolor{red}{\sf\bfseries Todo: #1}}

\newcommand{\red}[1]{\textcolor{red}{#1}}

\newcommand{\bfred}[1]{\textcolor{red}{\sf\bfseries #1}}

\newcommand{\blue}[1]{\textcolor{blue}{#1}}

\twocolumn[
\icmltitle{Lottery Ticket Preserves Weight Correlation: Is it Desirable or Not?}



\icmlsetsymbol{equal}{*}
\icmlsetsymbol{corresponding}{$\dagger$}

\begin{icmlauthorlist}
\icmlauthor{Ning Liu}{midea,equal}
\icmlauthor{Geng Yuan}{neu,equal}
\icmlauthor{Zhengping Che}{didi}
\icmlauthor{Xuan Shen}{neu}
\icmlauthor{Xiaolong Ma}{neu}
\icmlauthor{Qing Jin}{neu}
\icmlauthor{Jian Ren}{snap}
\icmlauthor{Jian Tang}{midea,corresponding}

\icmlauthor{Sijia Liu}{msu}
\icmlauthor{Yanzhi Wang}{neu,corresponding}

\end{icmlauthorlist}

\icmlaffiliation{midea}{Midea Group, Beijing, China}
\icmlaffiliation{neu}{Northeastern University, Boston, MA, USA}
\icmlaffiliation{didi}{Didi Chuxing, Beijing, China}
\icmlaffiliation{snap}{Snap Inc., CA, USA}
\icmlaffiliation{msu}{Michigan State University, MI, USA}



\icmlkeywords{Machine Learning, Deep Learning, Model Compression, ICML}

\vskip 0.3in
]



\printAffiliationsAndNotice{\icmlEqualContribution} 
\begin{abstract}

In deep model compression, the recent finding ``Lottery Ticket Hypothesis''~(LTH)~\cite{frankle2018lottery} pointed out that there could exist a  winning ticket (i.e., a properly pruned sub-network together with original weight initialization) that can achieve competitive performance than the original dense network.
However, it is not easy to observe such winning property in many scenarios, where for example, a relatively large learning rate is used even if it benefits training the original dense model.
In this work, we investigate the underlying condition and rationale behind the winning property, and find that the underlying reason is largely attributed to the correlation between initialized weights and final-trained weights when the learning rate is not sufficiently large. Thus, the existence of winning property is correlated with an insufficient DNN pretraining, and is unlikely to occur for a well-trained DNN. To overcome this limitation, we propose the ``pruning \& fine-tuning" method that consistently outperforms lottery ticket sparse training under the same pruning algorithm and the same total training epochs. Extensive experiments over multiple deep models (VGG, ResNet, MobileNet-v2) on different datasets have been conducted to justify our proposals.


\end{abstract}
\section{Introduction}

Weight pruning has been widely studied and utilized to effectively remove the redundant weights in the over-parameterized deep neural networks (DNNs) while maintaining the accuracy performance~\cite{han2015,han2016deep_compression,wen2016learning,he2017channel,min20182pfpce,he2019filter,dai2019nest,Lin_2020_CVPR,He_2020_CVPR}. The typical \emph{pruning pipeline} has three main stages: 1) train an over-parameterized DNN, 2) prune the unimportant weights in the original DNN, and 3) fine-tune the pruned DNN to restore accuracy.

Many works have been proposed to investigate the behaviors on weight pruning~\cite{SynFlow,ye2020good,renda2020comparing,malach2020proving}.
The Lottery Ticket Hypothesis (LTH)~\cite{frankle2018lottery} reveals that, inside a dense network with randomly initialized weights, a small sparse subnetwork, when trained in isolation using the identical initial weights, can reach a similar accuracy as the dense network. Such a sparse subnetwork with the initial weights is called the winning ticket. 

For a more rigorous definition, let $f(x;\theta_{0})$ be  a given network \emph{initialization}, where $\theta_{0}\sim D_{\theta}$ denotes the initial weights. We then formalize pre-training, pruning and sparse training below.
\emph{Pre-training:} The network is trained for $T$ epochs arriving at weights $\theta_T$ and network function $f(x;\theta_{T})$. \emph{Pruning:} Based on the pretrained weights $\theta_T$, adopt certain pruning algorithm to generate a sparse mask $m \in {\{0,1\}}^{|\theta|}$. \emph{Sparse Training:} The LTH paper considers two cases of sparse training. The first (``winning ticket'') is the direct application of mask $m$ to initial weights $\theta_{0}$, resulting in weights $\theta_{0}\odot m$ and network function $f(x;\theta_{0}\odot m)$. The second (random reinitialization) is the application of mask $m$ to a random initialization of weights $\theta_{0}^{\prime}\sim D_{\theta}$, resulting in weights $\theta'_{0}\odot m$ and network function $f(x;\theta'_{0}\odot m)$. The \emph{winning property} has \textbf{two} aspects \ding{172}-\ding{173} for identification: \ding{172} Training $f(x;\theta_{0}\odot m)$ for $T$ epochs (or fewer) will result in similar accuracy as that of the dense pre-trained network $f(x;\theta_{T})$. \ding{173} There should be a notable accuracy gap between training $f(x;\theta_{0}\odot m)$ for $T$ epochs and training $f(x;\theta'_{0}\odot m)$, and the former shall be higher.

In the standard LTH setup \cite{frankle2018lottery}, the winning property can be observed in the case of low learning rate via the simple iterative magnitude pruning algorithm, but fails to occur at higher initial learning rates especially in deeper neural networks.
For instance, the LTH work identifies the winning tickets on the CIFAR-10 dataset for the CONV-2/4/6 architectures (the down-scaled variants of VGG~\cite{simonyan2014very}), with the initial learning rate as low as 0.0001. For deeper networks such as ResNet-20 and VGG-19 on CIFAR-10, the winning tickets can be identified only in the case of low learning rate. At higher learning rates, additional warm up is needed to find the winning tickets.  In \citet{liu2018rethinking} (the latest ArXiv version), it revisits LTH and finds out that with a widely-adopted learning rate, the winning ticket has no accuracy advantage over the random reinitialization. This questions the second aspect of winning property on the accuracy gap between training $f(x;\theta_{0}\odot m)$ and training $f(x;\theta'_{0}\odot m)$.

Further, the following work~\citet{frankle2019stabilizing} proposes the iterative pruning with rewinding to stabilize identifying the winning tickets. Specifically, it resets the weights to $\theta_k$ in each pruning iteration, where $\theta_k$ denotes the weights trained from $\theta_0$ for a small number of $k$ epochs.

In this paper, we investigate the underlying condition and rationale behind winning property.
We ask if such a property is a natural characteristic of DNNs across their architectures and/or applications.
We revisit LTH via extensive experiments built upon various representative DNN models and datasets, and have confirmed that the winning property only exists at a low learning rate. 
In fact, such a ``low learning rate" (e.g., 0.01 for ResNet-20 and 0.0001 for CONV-2/4/6 architectures on CIFAR-10) is already significantly deviated from the standard learning rate, and results in notable accuracy degradation in the pretrained DNN. Besides, training from the ``winning ticket" at such a low learning rate can only restore the accuracy of the pretrained DNN under the same insufficient learning rate, instead of that under the desirable learning rate. By introducing a \emph{correlation indicator} for quantitative analysis, we found that the underlying reason is largely attributed to the correlation between initialized weights and final-trained weights when the learning rate is not sufficiently large. We draw the following conclusions:  
\begin{itemize}
    \item As a result of low learning rate, such weight correlation results in low accuracy in DNN pretraining.
    \item Such weight correlation is also a key condition of winning property, concluded through a detailed analysis of the cause of winning property.
    \item Thus, the existence of winning property is correlated with an insufficient DNN pretraining, i.e., it is unlikely to occur for a well-trained DNN.
\end{itemize}

Different from sparse training under lottery ticket setting, we propose the ``pruning \& fine-tuning" method, i.e., apply mask $m$ to pre-trained weights $\theta_T$ and perform fine-tuning for $T$ epochs. The generated sparse subnetwork can largely achieve the accuracy of the pretrained dense DNN.
Through comprehensive experiments and analysis we draw the following conclusions:
\begin{itemize}
\item ``Pruning \& fine-tuning" consistently outperforms lottery ticket setting under the same pruning algorithm for mask generation, and the same total training epochs.
\item The pruning algorithm responsible for mask generation plays an important role in the quality of generated sparse subnetwork.
    \item Thus, if one wants to optimize the accuracy of sparse subnetwork and restore the accuracy of the pretrained dense DNN, we suggest adopting the pruning \& fine-tuning method instead of lottery ticket setting.
\end{itemize}

\section{Related Work}
\subsection{DNN Weight Pruning}
DNN weight pruning as a model compression technique can effectively remove the redundant weights in DNN models and hence reduce both storage and computation costs.
The general flow of weight pruning consists of three steps: (1) train the neural network first; (2) derive a sub-network structure (i.e., removing unimportant weights) using a certain pruning algorithm; and (3) fine-tune the remaining weights in the sub-network to restore accuracy.
Different pruning algorithms will deliver different capabilities to search for the best-suited sparse sub-network and lead to different final accuracies.

The most straightforward method is the magnitude-based \emph{one-shot pruning}, which will directly zero-out a given percentage of trained weights with the smallest magnitude. 
However, this method usually leads to a severe accuracy drop under a relatively high pruning rate. 
\emph{Iterative magnitude pruning} is proposed in~\cite{han2015learning}, which removes the weights with the smallest magnitude in an iterative manner. It repeats step (1) and step (2) multiple times until reaching the target pruning rate. In~\cite{frankle2018lottery}, iterative pruning is adopted to find the sparse sub-network (i.e., winning ticket). The iterative pruning process is still a greedy search, and has been extended in~\cite{zhu2017prune,tan2020dropnet,Liu2020Autocompress} to derive better sub-network structures.

To overcome the greedy nature in the heuristic pruning methods, the more mathematics-oriented regularization-based algorithm~\cite{wen2016learning,he2017channel} has been proposed, to generate sparsity by incorporating $\ell_1$ or $\ell_2$ structured regularization in the loss function.
However, this method directly applies fixed regularization terms that penalize all weights equally and will lead to a potential accuracy drop.
Later work~\cite{zhang2018systematic,ren2019ADMMNN} incorporate Alternating Direction Methods of Multipliers (ADMM)~\cite{boyd2011distributed,ouyang2013stochastic} to solve the pruning problem as an optimization problem, which adopts dynamic regularization penalties and maintains high accuracy.

\subsection{Lottery Ticket Hypothesis}
\subsubsection{The Origin and Controversy of Lottery Ticket Hypothesis}
The recent work~\cite{frankle2018lottery} reveals that, inside a dense network with randomly initialized weights, a small sparse subnetwork can reach a similar test accuracy when trained in isolation using the identical initial weights as training the dense network. 
Such sparse subnetwork is called the winning ticket and can be found by pruning the pre-trained dense network under a non-trivial pruning ratio.


As demonstrated in~\cite{frankle2018lottery}, winning tickets can be found in small networks and small dataset when using relatively low learning rates (e.g., 0.01 for SGD). The work from the same period ~\cite{liu2018rethinking} finds that, when using a relatively large learning rate (e.g., 0.1 for SGD), training a ``winning ticket'' with identical initialized weights will not provide any unique advantage in accuracy compared to training with randomly initialized weights.
The following work~\cite{frankle2019stabilizing,renda2020comparing} also confirms that, for deeper networks and using relatively large learning rates, the winning property can hardly be observed. They propose a weight rewinding technique to identify small subnetworks, which can be trained in isolation to achieve competitive accuracy as the dense pretrained network.

\subsubsection{Other Aspects and Applications}

Later work~\cite{chen2020lottery} further extends the lottery ticket hypothesis to a pre-trained BERT model to evaluate the transferability of the sparse subnetworks among different downstream NLP tasks. Recent works~\cite{morcos2019one,chen2020lottery} have studied the lottery ticket hypothesis in computer vision tasks and in unsupervised learning.

The potential of sparse training suggested by the lottery ticket hypothesis has motivated the study of deriving the ``winning tickets'' at an early stage of training, thereby accelerating training process. There is a number of work in this direction~\cite{frankle2020pruning,you2019drawing,frankle2020early}, which are orthogonal to the discussions in this paper.




\section{Notations in this Paper}
\label{sec:notation}
\begin{figure}[htbp]
    \centering
    \includegraphics[width=0.9\columnwidth]{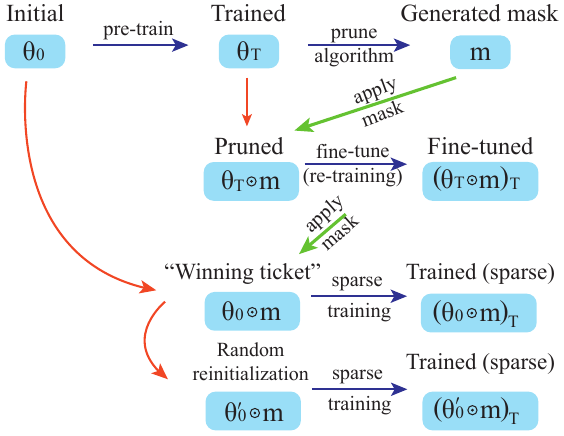}
    \caption{Illustration of notations: ``pre-training", ``pruning" (mask generation), ``sparse training", and ``pruning \& fine-tuning".}
    \label{fig:notation}
\end{figure}

In this paper, we follow the notations from \cite{frankle2018lottery} and generalize to the ``pruning \& fine-tuning" setup. Detailed notations (as shown in Figure~\ref{fig:notation}) are illustrated as follows:
\begin{itemize}
    \item \emph{Initialization}: Given a network $f(x;\theta_{0})$, where $\theta_{0}\sim D_{\theta}$ denotes the initial weights.
    \item \emph{Pre-training}: Train the network for $T$ epochs arriving at weights $\theta_T$ and network function $f(x;\theta_{T})$.
    \item \emph{Pruning}: Based on the trained weights $\theta_T$, adopt certain algorithm to generate a pruning mask $m \in {\{0,1\}}^{|\theta|}$. The LTH paper \cite{frankle2018lottery} uses the iterative pruning algorithm. We start from this algorithm for a fair evaluation, but are not restricted to it. Other algorithms, e.g., one-shot pruning and ADMM-based pruning are also employed to evaluate the impact on sparse training and pruning \& fine-tuning, as shown in  Section~\ref{sec:pruningfinetuninng}.
    \item \emph{Sparse Training (Lottery Ticket Setting):} The LTH paper considers two cases in the sparse training setup. The first is the direct application of mask $m$ to initial weights $\theta_{0}$, resulting in weights $\theta_{0}\odot m$ and network function $f(x;\theta_{0}\odot m)$. The LTH paper termed this case the ``winning tickets"\footnote{We inherit this terminology, although it does not result in the winning property in many of our testing results.}. The second is the application of mask $m$ to a random initialization of weights $\theta_{0}^{\prime}\sim D_{\theta}$, resulting in weights $\theta'_{0}\odot m$ (network function $f(x;\theta'_{0}\odot m)$). This case is termed ``random reinitialization" in the LTH paper. The weights after training $f(x;\theta_{0}\odot m)$ for $T$ epochs are denoted by $(\theta_{0}\odot m)_T$, while the weights after training $f(x;\theta'_{0}\odot m)$ for $T$ epochs are denoted by $(\theta'_{0}\odot m)_T$. Please note that the mask $m$ is kept through this training process.
    \item \emph{Pruning \& fine-tuning:} After generating the mask $m$, we directly apply it to the trained weights $\theta_T$, resulting in weights $\theta_{T}\odot m$, and perform fine-tuning (retraining) for another $T'$ epochs. The final weights are denoted by $(\theta_{T}\odot m)_{T'}$. To maintain the same number of total epochs as the lottery ticket setting, we set $T'=T$. Please note that the mask $m$ is kept through this fine-tuning process.
    \end{itemize}
The \emph{winning property} has twofold meaning: First, training $f(x;\theta_{0}\odot m)$ for $T$ epochs (or fewer) will result in similar accuracy as $f(x;\theta_{T})$ (pre-training result of the dense network), under a non-trivial pruning rate. Second, there should be a notable accuracy gap between training $f(x;\theta_{0}\odot m)$ for $T$ epochs and training $f(x;\theta'_{0}\odot m)$, and the former shall be higher.

\section{Why Lottery Ticket Exists? An Analysis from the Weight Correlation Perspective}
\label{sec:whyLT}

\subsection{Revisiting Lottery Ticket: When does this winning property exist?} 
\label{subsec:revisitLT}

We revisit the lottery ticket experiments on various DNN architectures including VGG-11, ResNet-20, and MobileNet-V2 on the CIFAR-10 and CIFAR-100 datasets. Our goal is to investigate the precise condition when winning property exists. We explore two different initial learning rates. The pruning approach for deriving masks follows the iterative pruning in \citet{frankle2018lottery}. Namely, iteratively remove a percentage of the weights with the least magnitudes in each layer. In each iterative pruning round, reset the weights to the initial weight $\theta_0$. We use the uniform per-layer pruning rate. Note the first convolutional layer is not pruned for all DNNs in this work. 

\begin{figure}[!h]
	\centering
	\begin{minipage}[b]{0.49\textwidth}
	\subfigure[Iterative pruning at learning rate of 0.01.]{
			\includegraphics[width=0.47\textwidth]{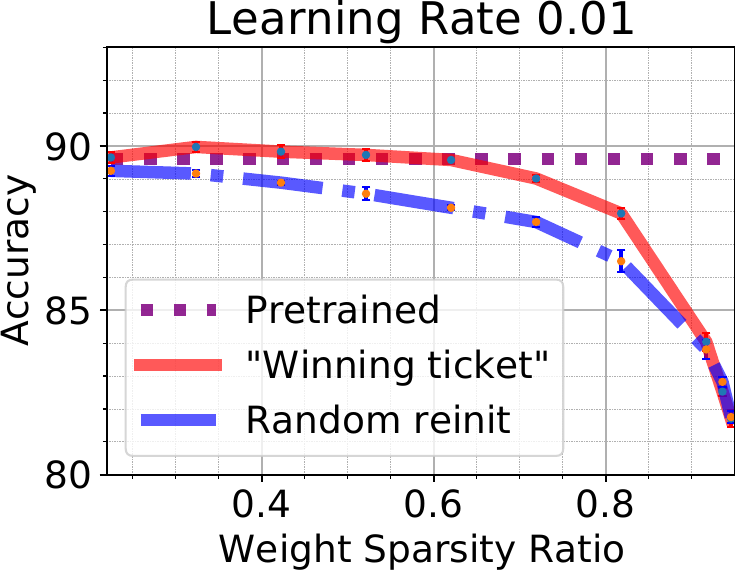}
			\label{fig:itlelr0.01}
	}
	\subfigure[Iterative pruning at learning rate of 0.1.]{
			\includegraphics[width=0.47\textwidth]{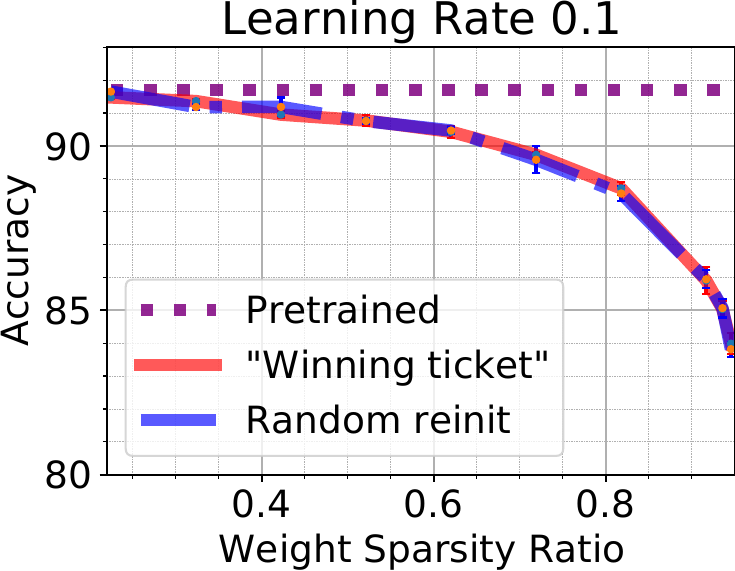}
			\label{fig:itlelr0.1}
	}
	\end{minipage}
	\caption{Illustration of random reinitialization and ``winning tickets'' for ResNet-20 on CIFAR-10 at learning rates 0.01 and 0.1.}
	\label{fig:iterativelottery}
\end{figure}

Figure~\ref{fig:iterativelottery} illustrates the experiments of accuracy comparison between random reinitialization and ``winning ticket'' (both sparse training) for ResNet-20 on CIFAR-10 at learning rates $0.01$ and $0.1$ over a range of different sparsity ratios (\citet{frankle2018lottery} uses the low learning rate 0.01). We conduct each experiment \textit{five} times (result variation shown in the figure). We set the same training epochs 150 rounds for training the original DNN with initial weights $f(x;\theta_{0})$ (i.e., pretraining), training from randomly reinitialized weights with the mask $f(x;\theta'_{0}\odot m)$ (random reinitialization), and training from the initial weights with the mask $f(x;\theta_0\odot m)$ (``winning ticket''). 
The hyperparameters used are the same as \cite{frankle2018lottery}: SGD with momentum (0.9), and the learning rates decrease by a factor of 10 after 80 and 120 epochs. The batch size is 128. No additional training tricks are utilized throughout the paper for fairness in comparison.

In the case of the initial learning rate of 0.01, the pre-trained DNN's accuracy is 89.62\%. The ``winning tickets'' consistently outperform the random reinitialization over different sparsity ratios. It achieves the highest accuracy 90.04\% (higher than the pre-trained DNN) at sparsity ratio of 62\%. This is similar to the observations found in \cite{frankle2018lottery} on the same network and dataset. On the other hand, in the case of the initial learning rate of 0.1, the pre-trained DNN's accuracy is 91.7\%. In this case, the ``winning ticket'' has a similar accuracy performance as the random reinitialization, and cannot achieve the accuracy close to the pre-trained DNN with a reasonable sparsity ratio (say 50\% or beyond). Thus no winning property is satisfied. Similar results can be found in the experiments of ResNet-20, VGG-11 and MobileNet-v2 on both CIFAR-10 and CIFAR-100, while the results of the rest of the experiments are detailed in Appendix~\ref{appendix:secA}.

From these experiments, the winning property exists at a low learning rate but does not exist at a relatively high learning rate, which is also observed in~\cite{liu2018rethinking}. However, we would like to point out that the relatively high learning rate 0.1 (which is in fact the standard learning rate on these datasets) results in a \textit{notably higher accuracy} in the pretrained DNN (91.7\% vs. 89.6\%) than the low learning rate\footnote{As CIFAR-10 is a relatively small dataset, 2\% accuracy is a notable accuracy difference that should not be ignored.}. The associated sparse training results (``winning ticket'', random reinitialization) in the lottery ticket setting are also higher with the learning rate 0.1. This point is largely missing in the previous discussions. Now the key \textbf{question} is: Are the above two observations correlated? If the answer is yes, it means that the winning property is not universal to DNNs, nor is it a natural characteristic of DNN architecture or application. Rather, it indicates that the learning rate is not sufficiently large, and the original, pretrained DNN is not well-trained.

Our hypothesis is that the above observations are correlated, and this is largely attributed to the \ul{correlation between initialized weights and final-trained weights when the learning rate is not sufficiently large}. Before validating our hypothesis, we will introduce a \emph{correlation indicator} (CI) for quantitative analysis.


\begin{figure}[htbp]
    \centering
    \includegraphics[width=0.98\columnwidth]{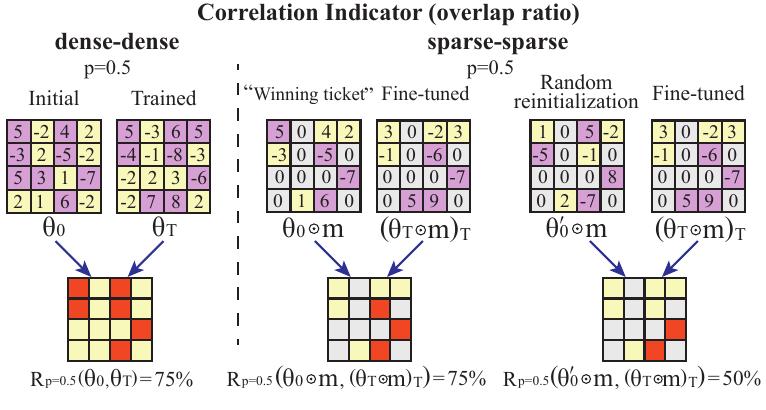}
    \caption{Scenarios for quantitative analysis of the weight correlation with an example of $sparsity\ ratio=50\%$ and $p=0.5$. This example is one DNN layer, while our actual computation is on the whole DNN.}
    \label{fig:correlation}
\end{figure}

\subsection{Weight Correlation Indicator}
\label{subsec:inndicator}
Consider a DNN with two collections of weights $\theta$ and $\theta'$. Note that this is a general definition that applies to both the original DNN and sparse DNN (when the mask is applied and a portion of weights eliminated). We define the \emph{correlation indicator} to quantify the amount of overlapped indices of large-magnitude weights between $\theta$ and $\theta'$. 
More specifically, given a DNN with $L$ layers, where the $l$-th layer has $N_l$ weights, the \emph{weight index set} $T_p\big((\theta)^l\big)$ ($p\in [0,1]$) is the top-$p\cdot 100\%$ largest-magnitude weights in the $l$-layer. Similarly, we define $T_p\big((\theta')^l\big)$. Please note that for a sparse DNN, the portion $p$ is defined with respect to the number of remaining weights in the sparse (sub-)network\footnote{In this way the formula can be unified for dense and sparse DNNs.}. The intersection of these two sets includes those weights that are large (top-$p\cdot 100\%$) in magnitude in both $\theta$ and $\theta'$, and $\textbf{card}\Big(T_p\big((\theta)^l\big)\cap T_p\big((\theta')^l\big)\Big)$ denotes the number of such weights in layer $l$. The correlation indicator (overlap ratio) between $\theta$ and $\theta'$ is finally defined as:
\begin{equation}
    R_p(\theta,\theta')=\frac{\sum_l\textbf{card}\Big(T_p\big((\theta)^l\big)\cap T_p\big((\theta')^l\big)\Big)}{p\cdot \sum_lN_l}
\end{equation}
When $R_p(\theta,\theta')\approx p$, the top-$p\cdot 100\%$ largest-magnitude weights in $\theta$ and $\theta'$ are largely independent. In this case the correlation is relatively weak\footnote{We cannot say that there is no correlation here because $R_p(\theta,\theta')\approx p$ is only a necessary condition.}. On the other hand, if there is a large deviation of $R_p(\theta,\theta')$ from $p$, then there is a strong correlation. Especially when $R_p(\theta,\theta')>p$, the weights that are large in magnitude in $\theta$ are likely to also be large in $\theta'$, indicating a positive correlation. Otherwise, when $R_p(\theta,\theta')<p$, it implies a negative correlation.

As shown in Figure~\ref{fig:correlation}, the above correlation indicator will be utilized to quantify the correlation between a dense DNN and a dense DNN, i.e., $R_p(\theta_0,\theta_T)$ for DNN pre-training, and between a sparse DNN and a sparse DNN, i.e., $R_p(\theta_{0}\odot m,(\theta_{T}\odot m)_{T})$ and $R_p(\theta'_{0}\odot m,(\theta_{T}\odot m)_{T})$ for the cases of ``winning ticket'' and random reinitialization under lottery ticket setting. Next, we will use the former to demonstrate the effect of different learning rates in DNN pre-training and the latter to demonstrate the rationale and condition of winning property.

\subsection{Weight Correlation in DNN Pre-Training}
\label{subsec:correlationpretrain}
Intuitively, the \emph{weight correlation} means that if a weight is large in magnitude at initialization, it is likely to be large after training. The reason for such correlation is that the learning rate is too low and weight updating is slow. Such weight correlation is not desirable for DNN training and typically results in lower accuracy, as weights in a well-trained DNN should depend more on the location of those weights instead of initialization~\cite{liu2018rethinking}. Thus when such weight correlation is strong, the DNN accuracy will be lower, i.e., not well-trained. 

To validate the above statement, we have performed experiments to derive $R_p(\theta_0,\theta_T)$ on DNN pretraining with different initial learning rates. Using ResNet-20 on CIFAR-10 dataset as an illustrative example. Figure 4 illustrates the correlation indicator between the initial weights $\theta_0$ and the trained weights $\theta_T$ from DNN pretraining at learning rates of 0.01 and 0.1, respectively. We use $T=150$ the same as Section~\ref{subsec:revisitLT}, also the same other hyperparameters and no additional training tricks.
We can observe that $R_p(\theta_0,\theta_T)$ at learning rate 0.01 has a notably higher correlation compared to the case of learning rate 0.1. This observation indicates that the large-magnitude weights of $\theta_0$ are not fully updated at a low learning rate 0.01, indicating that the pre-trained DNN is not well-trained. In the case of learning rate 0.1, the weights are sufficiently updated thus largely independent from the initial weights ($R_p(\theta_0,\theta_T)\approx p$, where $p=10\%, 20\%, 30\%, 40\%, 50\%$), indicating a well-trained DNN.
Results on other DNN models and datasets are provided in Appendix~\ref{appendix:secB}, and a similar conclusion can be drawn.

\begin{figure}[h!]
    \centering
    \includegraphics[width=0.98\columnwidth]{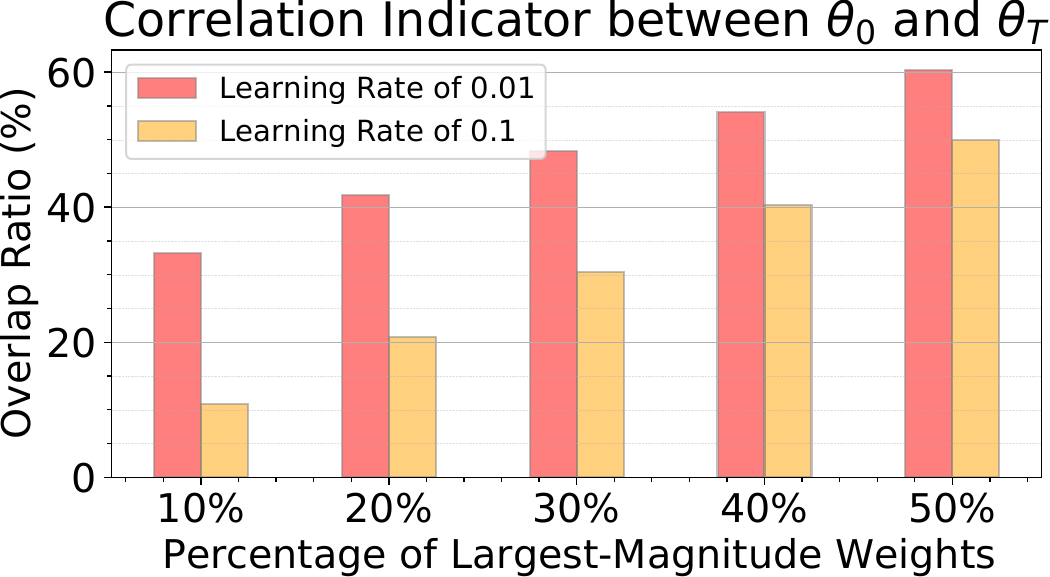}
    \caption{The overlap ratios (when $p=$ 10\%, 20\%, 30\%, 40\% and 50\%) between the initial weights $\theta_0$ and the pretrained weights $\theta_T$ at learning rate of 0.01 and 0.1.}
    \label{fig:overlapscratch}
\end{figure}

As shown in the result discussions, learning rates 0.1 and 0.01 (for ResNet-20) are not merely two candidate hyperparameter values. Rather, they result in a well-trained DNN (so a desirable learning rate) and a not well-trained DNN (so a not-so-good learning rate), respectively. We shall not rely on the conclusion drawn from the latter that results in an insufficient DNN pre-training.

\subsection{Cause and Condition of the Winning Property}
\label{subsec:causeLTH}

\noindent\textbf{Weight Correlation under Lottery Ticket Setting:} In this subsection, our goal is to understand the different accuracy from training $f(x;\theta_{0}\odot m)$ (``winning ticket'') and training $f(x;\theta'_{0}\odot m)$ (random reinitialization) when the learning rate is low, thereby revealing the cause and condition of winning property. We will achieve this goal by studying the weight correlation. 

\begin{figure}[h!]
	\centering
	\begin{minipage}[b]{0.49\textwidth}
	\subfigure[Pruning\&fine-tuning at learning rate of 0.01.]{
		\includegraphics[width=0.47\textwidth]{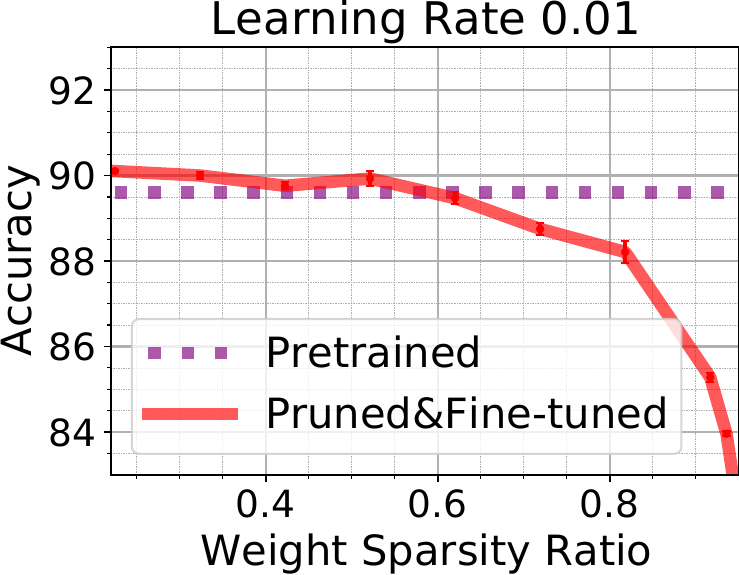}
		\label{fig:sparsefinetune0.01}
	}
	\subfigure[Pruning\&fine-tuning at learning rate of 0.1.]{
		\includegraphics[width=0.47\textwidth]{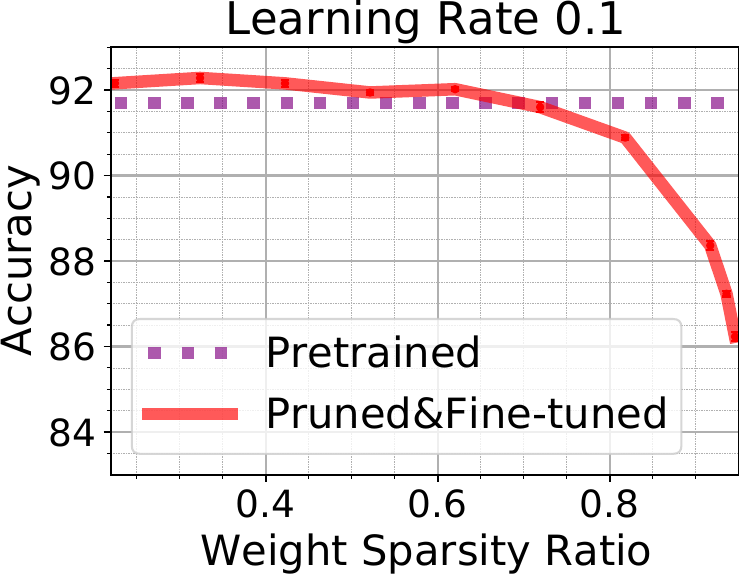}
		\label{fig:sparsefinetune0.1}
	}
	\end{minipage}
	
	\begin{minipage}[b]{0.49\textwidth}
	\subfigure[The overlap ratio comparison at learning rate 0.01.]{
		\includegraphics[width=0.47\textwidth]{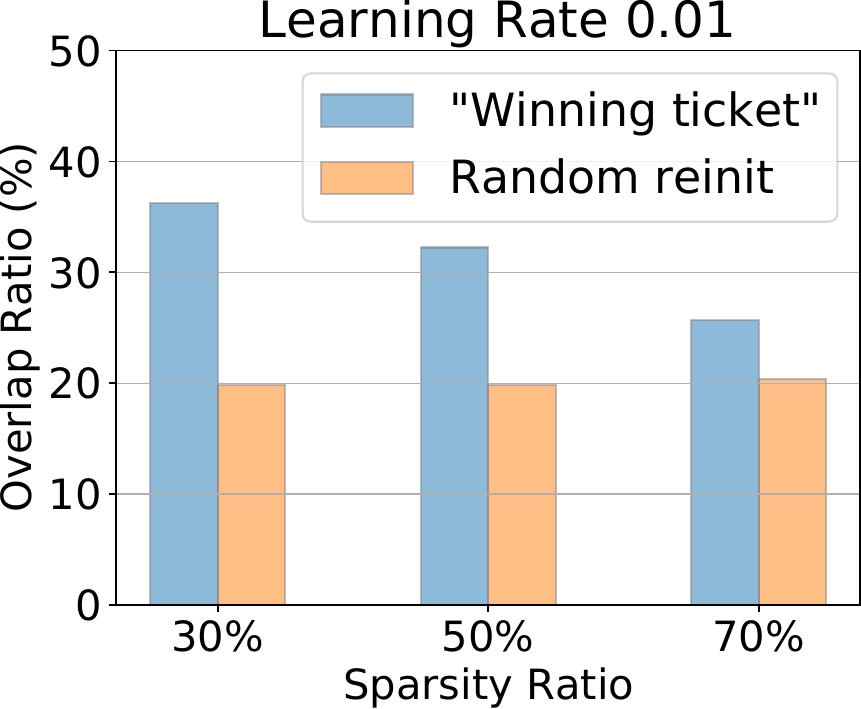}
		\label{fig:Overlapsparse0.01}
	}
	\subfigure[The overlap ratio comparison at learning rate 0.1.]{
		\includegraphics[width=0.47\textwidth]{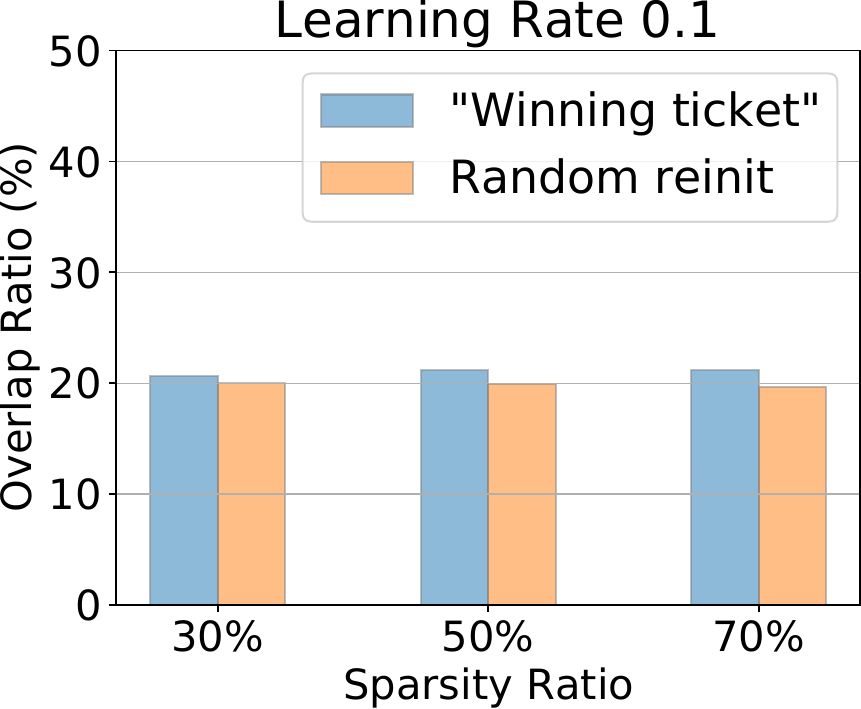}
		\label{fig:Overlapsparse0.1}
	}
	\end{minipage}
	
	\caption{(a), (b): Accuracies of $f\big(x;(\theta_{T}\odot m)_{T}\big)$ (``pruning \& fine-tuning") at different sparsity ratios with masks generated by iterative pruning. (c), (d): The weight correlation (overlap ratio) comparison at $p=0.2$, between $\theta_0\odot m$ (``winning ticket'') and $(\theta_T\odot m)_T$ (pruned\&fine-tuned weights), and between $\theta_0^{\prime}\odot m$ (random reinitialization) and $(\theta_T\odot m)_T$ (pruned\&fine-tuned weights) under 0.3, 0.5, 0.7 sparsity ratios.}
	\label{fig:sparseoverlap}
\end{figure}

Consider the ``pruning \& fine-tuning" case formally defined in Section~\ref{sec:notation}, in which we apply mask $m$ on the trained weights $\theta_T$ from DNN pretraining, and then perform fine-tuning for another $T$ epochs. The final weights are denoted by $(\theta_{T}\odot m)_{T}$. Using ResNet-20 on CIFAR-10 as an illustrative example. Figure~\ref{fig:sparsefinetune0.01} and~\ref{fig:sparsefinetune0.1} show the accuracy of the ``pruning \& fine-tuning" result $f\big(x;(\theta_{T}\odot m)_{T}\big)$ at different sparsity ratios, with learning rates 0.01 and 0.1, respectively. Again we use $T=150$ epochs and the same other hyperparameters. The accuracies of the pretrained DNN with corresponding learning rates are also provided. One can observe that $f\big(x;(\theta_{T}\odot m)_{T}\big)$ achieves relatively high accuracy, close to or higher than the accuracy of the pretrained DNN at the same learning rate (even at the desirable learning rate 0.1)\footnote{In fact, the relatively high accuracy of $f\big(x;(\theta_{T}\odot m)_{T}\big)$ is one major reason for us to explore the correlation between $\theta_{0}\odot m$ ($\theta'_{0}\odot m$) and $(\theta_{T}\odot m)_{T}$. In Section~\ref{sec:pruningfinetuninng} we will generalize to the conclusion that ``pruning \& fine-tuning" results in higher accuracy in general than sparse training (the lottery ticket setting).}. Results on other DNN models and datasets are provided in Appendix~\ref{appendix:secC}, and a similar conclusion can be drawn.

We study the correlation between $\theta_{0}\odot m$ ($\theta'_{0}\odot m$) and $(\theta_{T}\odot m)_{T}$ to shed some light on the cause of winning property. Again use ResNet-20 on CIFAR-10 as an illustrative example, while the results on other DNN models and datasets are provided in Appendix~\ref{appendix:secD} with similar conclusion. Figure~\ref{fig:Overlapsparse0.01} shows the correlation indicator between $\theta_{0}\odot m$ (``winning ticket'') and $(\theta_{T}\odot m)_{T}$, and between $\theta'_{0}\odot m$ (random reinitialization) and $(\theta_{T}\odot m)_{T}$, under the insufficient learning rate 0.01. While Figure~\ref{fig:Overlapsparse0.1} shows the correlation indicator results under the desirable learning rate 0.1. One can observe the positive correlation between $\theta_{0}\odot m$ and $(\theta_{T}\odot m)_{T}$ at the low learning rate, when the winning property exists. Such correlation is minor in the other cases.

\noindent\textbf{Analysis of Weight Correlation and Condition of Winning Property:} Let us investigate the \textbf{cause} of correlation between $\theta_{0}\odot m$ and $(\theta_{T}\odot m)_{T}$ at low learning rate. As shown in Section~\ref{subsec:correlationpretrain}, there is a correlation between $\theta_0$ and $\theta_T$ at the insufficient learning rate. Then there is also a correlation between $\theta_0\odot m$ and $\theta_T\odot m$ (both applied the same mask). As $\theta_T\odot m$ includes the pretrained weights and $(\theta_{T}\odot m)_{T}$ only applies additional fine-tuning, there will be positive correlation between $\theta_T\odot m$ and $(\theta_{T}\odot m)_{T}$. Combining the above two statements will yield the correlation between $\theta_{0}\odot m$ and $(\theta_{T}\odot m)_{T}$. When we consider random reinitialization, there is no correlation between $\theta'_0$ and $\theta_T$ as a reinitialization is applied. So there is no correlation between $\theta'_0\odot m$ and $\theta_T\odot m$, or between $\theta'_{0}\odot m$ and $(\theta_{T}\odot m)_{T}$.

At a desirable learning rate 0.1, there is a minor (or no) correlation between $\theta_0$ and $\theta_T$ as shown in Section~\ref{subsec:correlationpretrain}. As a result, there is minor (or no) correlation between $\theta_0\odot m$ and $\theta_T\odot m$, or between $\theta_{0}\odot m$ and $(\theta_{T}\odot m)_{T}$. From the above analysis, one can observe that the correlation between $\theta_0$ and $\theta_T$ is the \textbf{key condition} in weight correlation analysis. 

The positive correlation between $\theta_{0}\odot m$ and $(\theta_{T}\odot m)_{T}$ helps to explain the winning property at low learning rate. Compared with random reinitialization $\theta'_{0}\odot m$, the ``winning ticket'' $\theta_{0}\odot m$ is ``closer" to a reasonably accurate solution. As the weight upscaling is slow (learning rate insufficient), it takes less effort to reach a higher accuracy starting from $\theta_{0}\odot m$ compared with starting from $\theta'_{0}\odot m$. Besides, as pointed out in Section~\ref{subsec:correlationpretrain}, the insufficient learning rate (and correlation between $\theta_0$ and $\theta_T$) results in a low accuracy in the pre-trained DNN, which makes it easier for sparse training to reach its accuracy. On the other hand, at a sufficient learning rate, such correlations do not exist (or are very minor), and then the winning property does not exist.

\noindent\textbf{Remarks:} From the above analysis, we conclude that a key condition of winning property is the correlation between $\theta_0$ and $\theta_T$. However, as already demonstrated in Section~\ref{subsec:correlationpretrain}, under the same condition the pretrained network will not be well-trained, as weights in a well-trained DNN should depend more on the location of those weights instead of initialization. In fact, as shown in Figure~\ref{fig:iterativelottery} and Appendix~\ref{appendix:secA}, the ``winning ticket'' can only restore the accuracy of the pretrained DNN under the same insufficient learning rate, instead of reaching the pretrained DNN accuracy at a desirable learning rate. This makes the value of the winning property questionable.

\subsection{Takeaway}
As discussed above, the existence of winning property is correlated with an insufficient DNN pretraining. It seems that winning property is not a natural characteristic of DNN architecture or application, and is unlikely to occur for a well-trained DNN (with a desirable learning rate). As a result, we do not suggest investigating the winning property under an insufficient learning rate.

\section{Pruning \& Fine-tuning -- A Better Way to Restore Accuracy under Sparsity}
\label{sec:pruningfinetuninng}

As concluded from the above discussions, it is difficult for sparse training to restore the accuracy of the pre-trained dense DNN, when a desirable learning rate is applied. 
On the other hand, as already hinted in Section~\ref{subsec:causeLTH}, the ``pruning \& fine-tuning" (i.e., fine-tuning from $(\theta_{T}\odot m)_{T}$) exhibits a higher capability in achieving the accuracy of a pre-trained DNN, no matter what the learning rate is.
Compared with the lottery ticket setting, the only difference in ``pruning \& fine-tuning" is that the mask $m$ is applied to the pretrained weights $\theta_T$. Is this the key reason for the high accuracy? Is this a universal property for different DNN architectures and applications? If the answer is yes, what is the underlying reason? We aim to answer these questions.

In this section, we only consider the \textbf{desirable learning rate} (e.g., 0.1 for ResNet-20 on CIFAR-10 dataset) and sufficient DNN pre-training, as the associated conclusions will be more meaningful.

\textbf{Fair Comparison with Lottery Ticket Setting:} We claim that under the same $T$ epochs for fine-tuning and sparse training, it is a fair comparison. The generation of mask $m$ is the same. The only difference is that pruning \& fine-tuning applies mask $m$ to the pre-trained weights $\theta_T$, while sparse training applies $m$ to $\theta_0$. Note that $\theta_T$ is available before $m$ as the latter is derived based on $\theta_T$ using pruning algorithm. Thus pruning \& fine-tuning will have \textbf{no additional training epochs} compared with sparse training.

\begin{figure*}[!h]
	\centering
	\begin{minipage}[b]{0.95\textwidth}
	\subfigure[The iterative pruning at learning rate of 0.1.]{
		\includegraphics[width=0.32\textwidth]{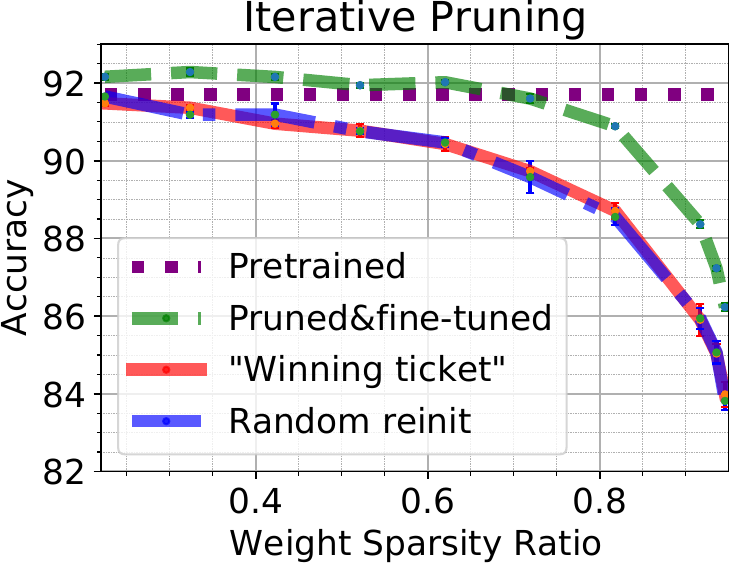}
		\label{fig:iterativefinetune0.1}
	}
	\subfigure[The ADMM-based pruning at learning rate of 0.1.]{
		\includegraphics[width=0.32\textwidth]{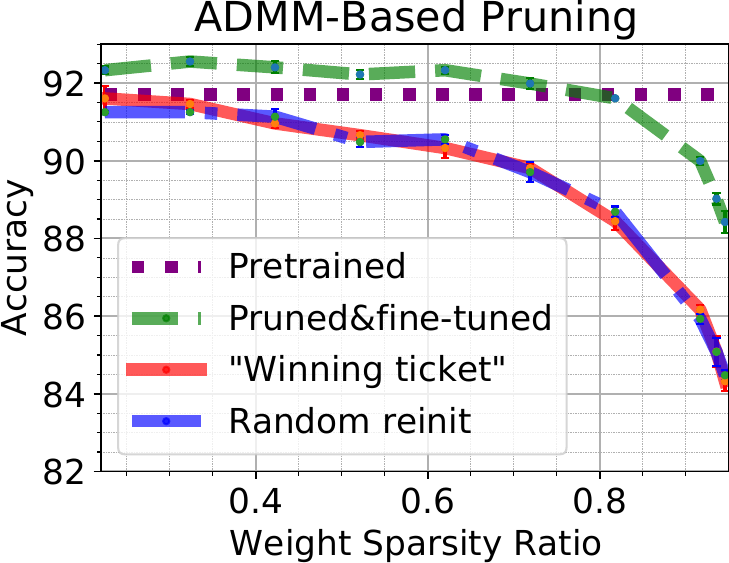}
		\label{fig:admmfinetune0.1}
	}
	\subfigure[The least magnitude one-shot pruning at learning rate of 0.1.]{
		\includegraphics[width=0.32\textwidth]{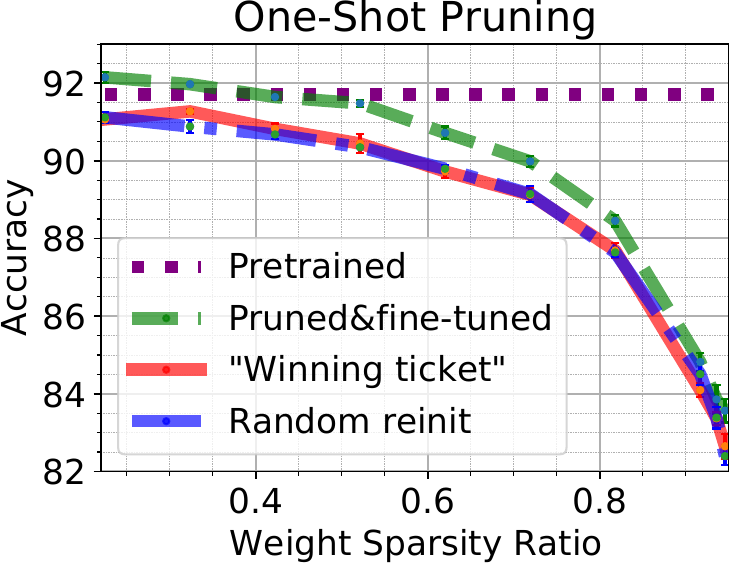}
		\label{fig:oneshotfinnetune0.1}
	}
	\end{minipage}
	\vspace{-2mm}
	\caption{Accuracy performance of pruning \& fine-tuning vs. two sparse training cases (``winning ticket'' and random reinitialization). Three pruning algorithms are utilized for mask generation: iterative pruning (a), ADMM-based pruning \cite{zhang2018systematic} (b), and one-shot pruning (c).}
	\label{fig:prunefinetune}
	\vspace{-2mm}
\end{figure*}

\begin{figure}[h!]
    \centering
    \includegraphics[width=0.85\columnwidth]{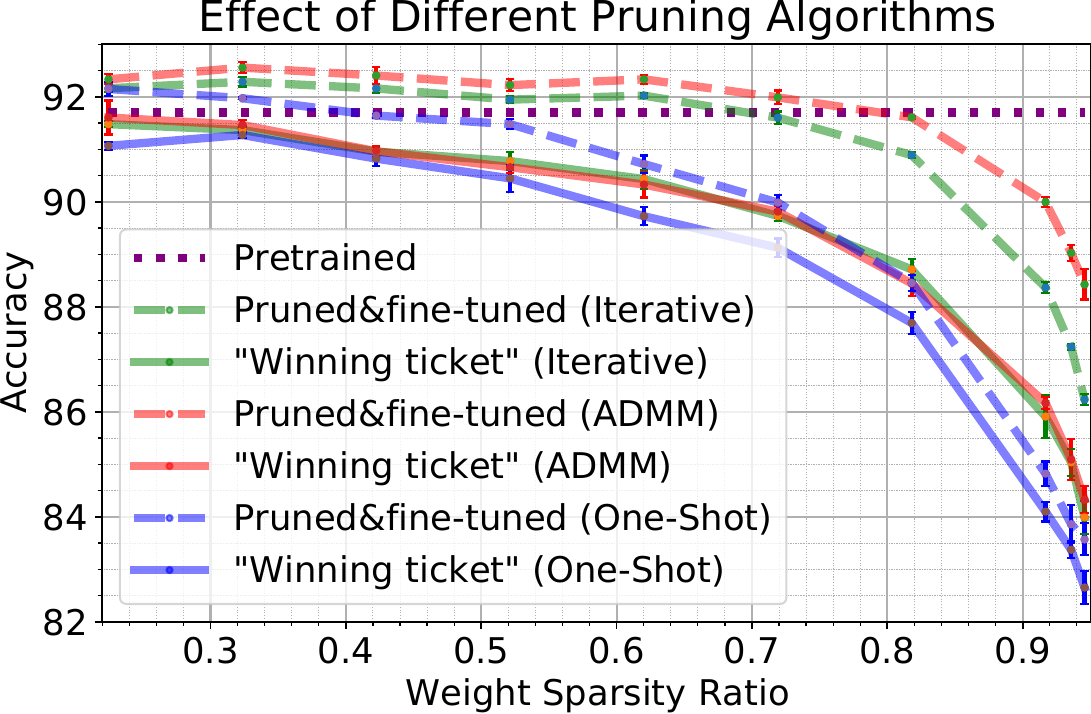}
    \vspace{-2mm}
    \caption{Accuracy performances of pruning \& fine-tuning and sparse training (``winning ticket'' case), under all three pruning algorithms (iterative pruning, ADMM-based pruning, and one-shot pruning) for mask generation.}
    \label{fig:prunefinetunevsticket}
    \vspace{-2mm}
\end{figure}

\textbf{Comparison between Pruning \& Fine-tuning and Sparse Training:} We use ResNet-20 on CIFAR-10 dataset as an illustrative example, and the rest of results are provided in Appendix~\ref{appendix:secE} (with the similar conclusion). We use the desirable learning rate 0.1, $T=150$ epochs, and the same as Section~\ref{subsec:revisitLT} for the rest of hyperparameters. Figure~\ref{fig:iterativefinetune0.1} shows the accuracy comparison between pruning \& fine-tuning (i.e., training (fine-tuning) from $\theta_{T}\odot m$) and the two sparse training scenarios ``winning ticket'' (i.e., training from $\theta_{0}\odot m$) and random reinitialization (i.e., training from $\theta'_{0}\odot m$) at different sparsity ratios. Iterative pruning algorithm is used to derive mask $m$ here. One can clearly observe the accuracy gap between pruning \& fine-tuning and the two sparse training cases (lottery ticket setting). In fact, the pruning \& fine-tuning scheme can consistently outperform the pretrained dense DNN up to sparsity ratio 70\%. Again, there is no accuracy difference between the two sparse training cases. 

Furthermore, we consider other two candidate pruning algorithms, ADMM-based pruning \cite{zhang2018systematic} and one-shot pruning, for pruning mask generation. Figure \ref{fig:admmfinetune0.1} and Figure \ref{fig:oneshotfinnetune0.1} demonstrate the corresponding accuracy comparison results between pruning \& fine-tuning and the two sparsity training scenarios. Again one can observe the notable advantage of pruning \& fine-tuning over the lottery ticket setting, even with a weak one-shot pruning algorithm for mask generation. In fact, pruning \& fine-tuning under ADMM-based pruning can restore the accuracy of pretrained DNN with 80\% sparsity. The property is not found under any of these pruning algorithms. Clearly, the consistent advantage of pruning \& fine-tuning is attributed to the fact that mask $m$ is applied to pretrained weights $\theta_T$ instead of the initialized weights $\theta_0$. In fact, information in $\theta_T$ is important for the sparse subnetwork to maintain accuracy of the pretrained dense network. Or in other words, weights in the desirable sparse subnetwork should have correlation with $\theta_T$ instead of $\theta_0$.

\textbf{Effect of Different Pruning Algorithms -- Towards a Better Mask Generation:} We have tested three pruning algorithms for mask generation. How to evaluate their relative performance? Figure~\ref{fig:prunefinetunevsticket} combines the above results and demonstrates the accuracy performances of pruning \& fine-tuning and sparse training (``winning ticket'' case), under all three pruning algorithms. The rest of results are in Appendix~\ref{appendix:secE}. One can observe the order in the accuracy performance: ADMM-based pruning on top, iterative pruning in the middle, and one-shot pruning the lowest. This order is the same for pruning \& fine-tuning and sparse training.
Note that the pruning algorithm is utilized to generate mask $m$, while the other conditions are the same (i.e., $\theta_T$, fine-tuning $T$ epochs on $\theta_T\odot m$, or sparse training on $\theta_0\odot m$). Hence, the relative performance is attributed to the quality in mask generation. One can conclude that the selection of pruning algorithm is critical in generating the sparse subnetwork as the quality of mask generation plays a key role here.

\textbf{An Analysis from Weight Correlation Perspective:} We conduct a weight correlation analysis of pruning \& fine-tuning results that can largely restore the accuracy of pretrained, dense DNN, between the final weights $(\theta_{T}\odot m)_{T}$ and the initialization $\theta_0$. Detailed results and discussions are provided in Appendix~\ref{appendix:secF}. The major conclusion is that there is a lack of correlation between $(\theta_{T}\odot m)_{T}$ and $\theta_0$, but there is a correlation between $(\theta_{T}\odot m)_{T}$ and $\theta_T$. It further strengthens the conclusion that it is not desirable to have the weight correlation between final-trained weights and weight initialization.

\textbf{Comparison with \citet{frankle2019stabilizing}:} The work \citet{frankle2019stabilizing} suggests applying mask $m$ to $\theta_k$ and then apply sparse training, where $\theta_k$ denotes the weights trained from $\theta_0$ for a small number of $k$ epochs. This technique is training from $\theta_{k}\odot m$, and is in between sparse training (training from $\theta_{0}\odot m$) and pruning \& fine-tuning (training from $\theta_{T}\odot m$). We point out that these three cases require \textbf{the same number of total epochs} under the same pruning algorithm, as mask $m$ is generated later than $\theta_k$ or $\theta_T$. We conduct a comprehensive comparison on the relative performance, with detailed results and discussions in Appendix~\ref{appendix:secG}. The major conclusion is that pruning \& fine-tuning consistently outperforms the method \citet{frankle2019stabilizing} over different sparsity ratios, DNN models, and datasets. As they exhibit the same number of training epochs, we suggest directly applying the mask $m$ to $\theta_T$ and perform fine-tuning, instead of applying to $\theta_k$.

\textbf{Remarks:} If one wants to optimize the accuracy of sparse subnetwork and restore the accuracy of the pretrained dense DNN, we suggest adopting the pruning \& fine-tuning method instead of lottery ticket setting.

\section{Conclusion}
In this work, we investigate the underlying condition and rationale behind lottery ticket property. We introduce a correlation indicator for quantitative analysis. Extensive experiments over multiple deep models on different datasets have been conducted to justify that the existence of winning property is correlated with an insufficient DNN pretraining, and is unlikely to occur for a well-trained DNN. Meanwhile, the sparse training of lottery ticket setting is difficult to restore the accuracy of the pre-trained dense DNN. To overcome this limitation, we propose the ``pruning \& fine-tuning'' method that consistently outperforms lottery ticket sparse training under the same pruning algorithm and total training epochs over various DNNs on different datasets.

\section*{Acknowledgements}
This research was supported by the National Science Foundation under awards CNS-1704662 and CCF-1937500.
Any opinions, findings, and conclusions or recommendations  in this material are those of the authors and do not necessarily reflect the views of NSF. 



\bibliography{reference}
\bibliographystyle{icml2021}

\appendix

\setcounter{equation}{0}
\setcounter{figure}{0}
\setcounter{table}{0}


\clearpage
{\huge \bf Appendix}

We explore a variety of deep neural networks (DNNs) to support our results. All experiments on CIFAR-10 and CIFAR-100 use the basic data augmentation including random cropping and random horizontal flipping.
No other tricks are used. We use 150 training epochs ($T=150$) and decay the initial learning rate after 80, 120 epochs by a factor of 10 (with step learning rate scheduler). The optimizer is stochastic gradient descent (SGD) and the momentum is 0.9. This training setup is used for pre-training ($f(x;\theta_T)$), pruning \& fine-tuning ($(\theta_T\odot m)_T$), and sparse training ($(\theta_0\odot m)_T$) as introduced in Section~\ref{sec:notation}.

\section{Revisit Lottery Tickets}
\label{appendix:secA}

We show the experiment results of MobileNet-V2 on CIFAR-10, and ResNet-20, VGG-11, and MobileNet-V2 on CIFAR-100 over a range of different sparsity ratios with the masks generated from iterative pruning~\cite{frankle2018lottery} at learning rate 0.01 and 0.1, respectively. 
We conduct each experiment \textit{five} times (result variation shown in figures). We set the same training epochs (i.e., 150 epochs) for training the original DNNs with initial weights $f(x;\theta_{0})$ (i.e., pretraining), training from randomly reinitialized weights with the mask $f(x;\theta'_{0}\odot m)$ (random reinitialization), and training from the initial weights with the mask $f(x;\theta_0\odot m)$ (``winning ticket'').

\begin{figure}[t!]
	\centering
	\begin{minipage}[b]{0.95\columnwidth}
	\subfigure[Iterative pruning at learning rate of 0.01 on MobileNet-V2 using CIFAR-10.]{
		\includegraphics[width=0.47\columnwidth]{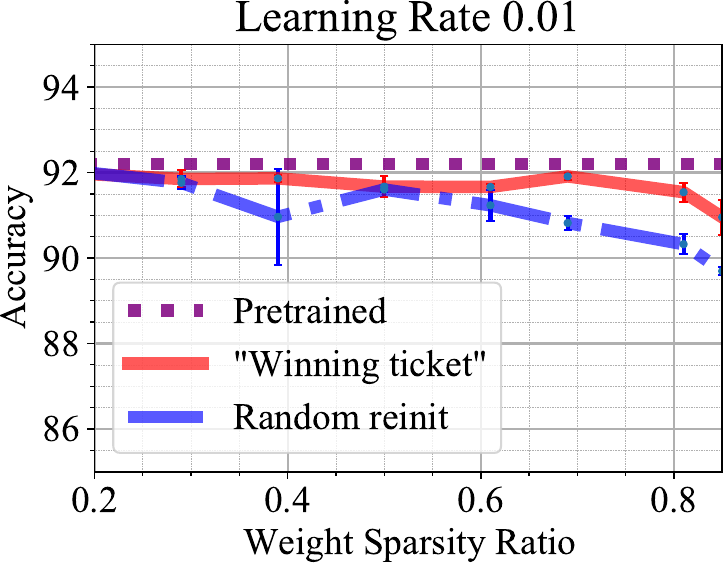}
		\label{appendixfig:fig2_mobilenet_0.01_cifar10}
	}
	\hfill
	\subfigure[Iterative pruning at learning rate of 0.1 on Mobilenet-V2 using CIFAR-10.]{
		\includegraphics[width=0.47\columnwidth]{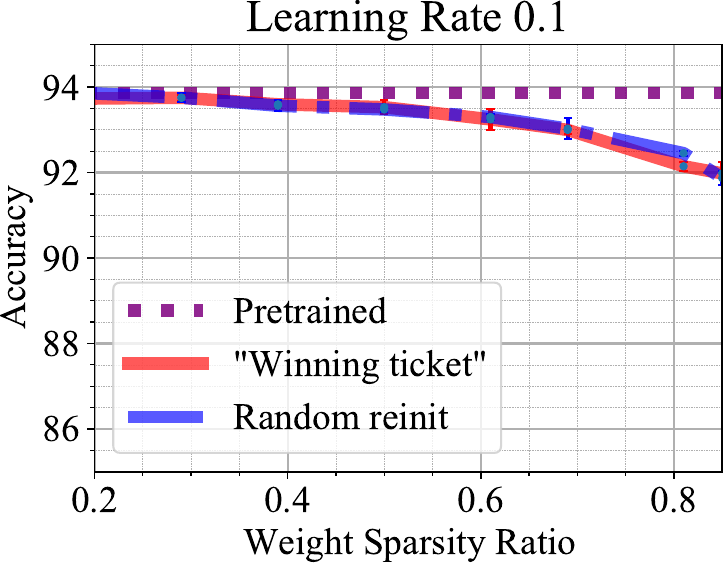}
		\label{appendixfig:fig2_mobilenet_0.1_cifar10}
	}
	\subfigure[Iterative pruning at learning rate of 0.01 on MobileNet-V2 using CIFAR-100.]{
		\includegraphics[width=0.47\columnwidth]{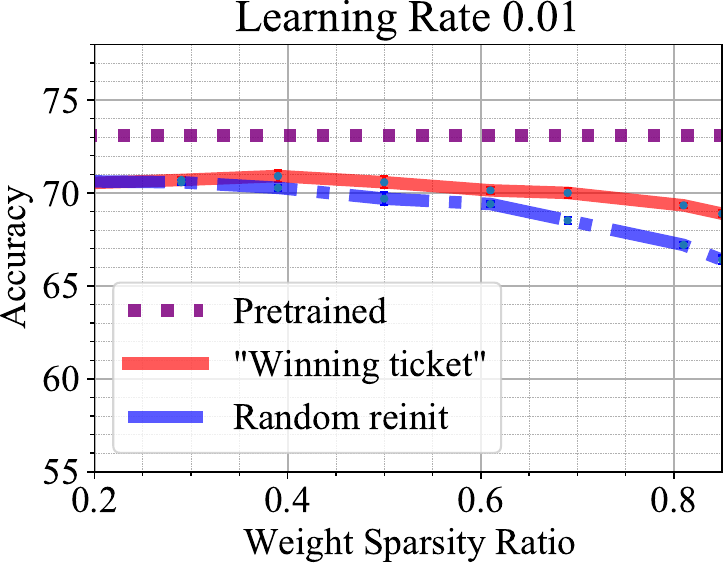}
		\label{appendixfig:fig2_mobilenet_0.01_cifar100}
	}
	\hfill
	\subfigure[Iterative pruning at learning rate of 0.1 on MobileNet-V2 using CIFAR-100.]{
		\includegraphics[width=0.47\columnwidth]{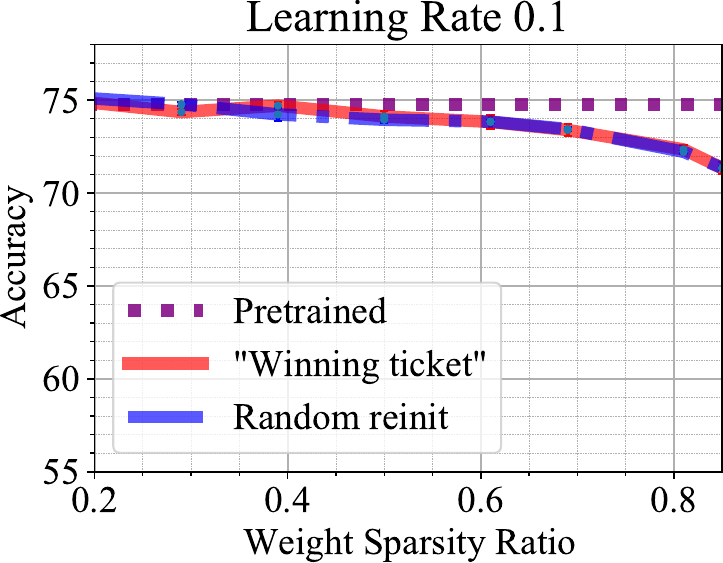}
		\label{appendixfig:fig2_mobilenet_0.1_cifar100}
	}
	\subfigure[Iterative pruning at learning rate of 0.01 on ResNet-20 using CIFAR-100.]{
		\includegraphics[width=0.47\columnwidth]{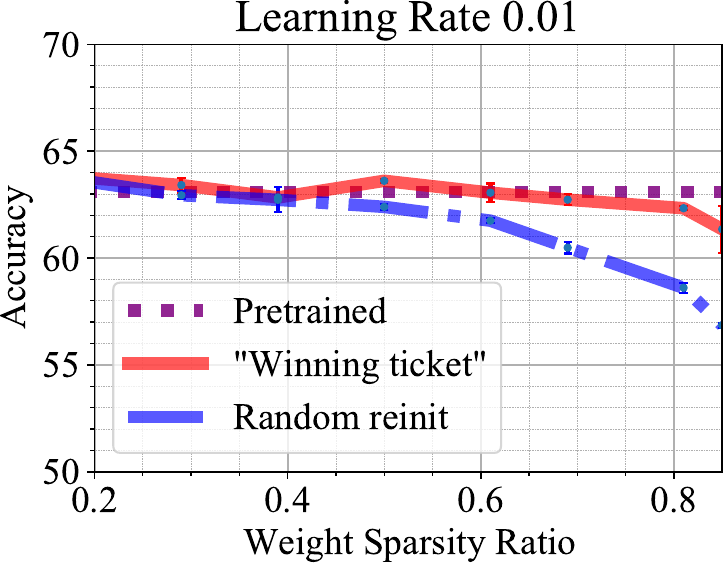}
		\label{appendixfig:fig2_resnet20_0.01_cifar100}
	}
	\hfill
	\subfigure[Iterative pruning at learning rate of 0.1 on ResNet-20 using CIFAR-100.]{
		\includegraphics[width=0.47\columnwidth]{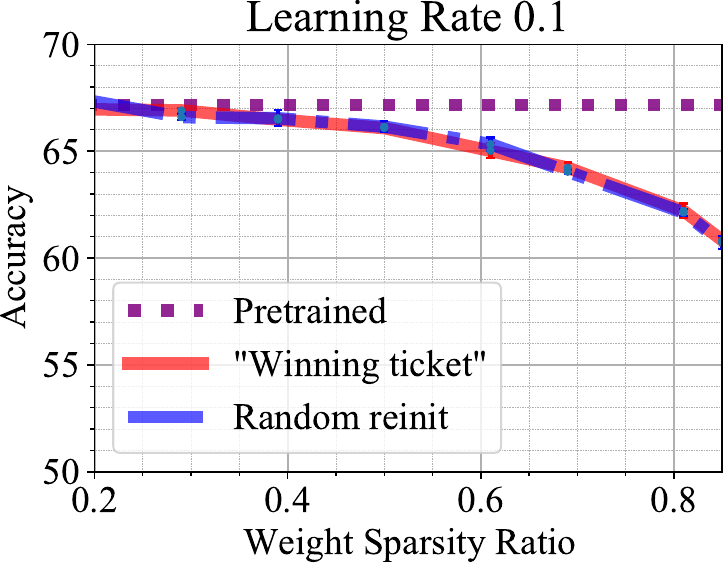}
		\label{appendixfig:fig2_resnet20_0.1_cifar100}
	}
	\subfigure[Iterative pruning at learning rate of 0.01 on VGG-11 using CIFAR-100.]{
		\includegraphics[width=0.47\columnwidth]{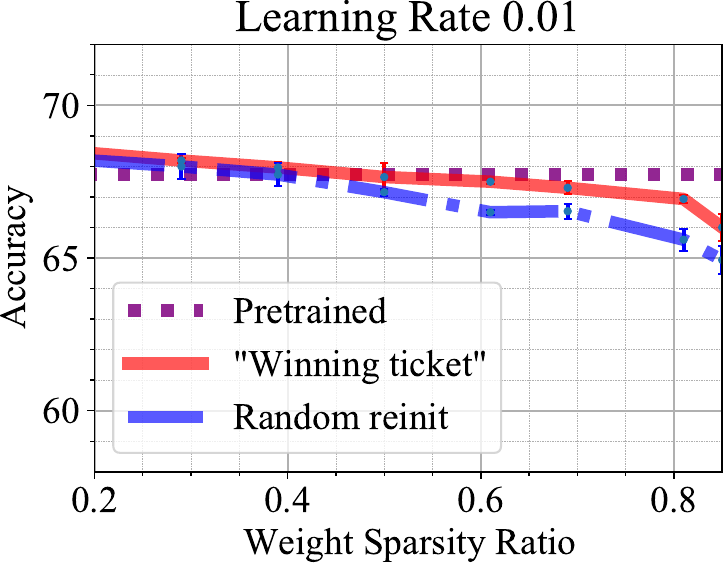}
		\label{appendixfig:fig2_vgg11_0.01_cifar100}
	}
	\hfill
	\subfigure[Iterative pruning at learning rate of 0.1 on VGG-11 using CIFAR-100.]{
		\includegraphics[width=0.47\columnwidth]{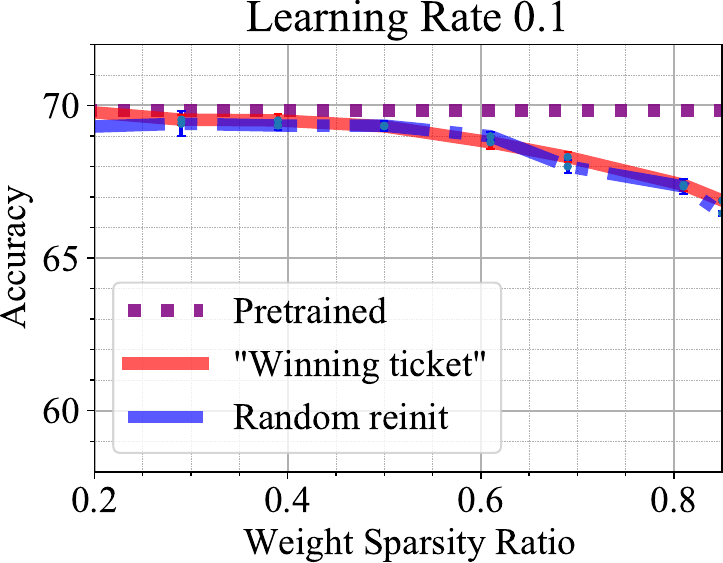}
		\label{appendixfig:fig2_vgg11_0.1_cifar100}
	}
	
	\end{minipage}
	\caption{Accuracy illustration of random reinitialization and ``winning tickets'' for MobileNet-V2 on CIFAR-10, and MobileNet-V2, ResNet-20 and VGG-11 on CIFAR-100 at learning rates 0.01 and 0.1.}
	\label{appendixfig:overlapscratch}
\end{figure}

\textbf{CIFAR-10 Results:} Figure~\ref{appendixfig:fig2_mobilenet_0.01_cifar10} and~\ref{appendixfig:fig2_mobilenet_0.1_cifar10} illustrate the result on MobileNet-V2 using CIFAR-10. The pre-trained MobileNet-V2’s accuracy on CIFAR-10 is 92.20\% at initial learning rate 0.01, and 93.86\% at initial learning rate 0.1.

\textbf{CIFAR-100 Results:} Figure~\ref{appendixfig:fig2_mobilenet_0.01_cifar100} and~\ref{appendixfig:fig2_mobilenet_0.1_cifar100} show the result on MobileNet-V2 using CIFAR-100. The pre-trained MobileNet-V2’s accuracy on CIFAR-100 is 73.10\% at initial learning rate 0.01, and 74.76\% at initial learning rate 0.1.
Figure~\ref{appendixfig:fig2_resnet20_0.01_cifar100} and ~\ref{appendixfig:fig2_resnet20_0.1_cifar100} show the result on ResNet-20 for CIFAR-100. The pre-trained ResNet-20’s accuracy on CIFAR-100 is 63.10\% at initial learning rate 0.01, and 67.15\% at initial learning rate 0.1 (see the significant gap here). 
Figure~\ref{appendixfig:fig2_vgg11_0.01_cifar100} and ~\ref{appendixfig:fig2_vgg11_0.1_cifar100} show the result on VGG-11 for CIFAR-100. The pre-trained VGG-11’s accuracy on CIFAR-100 is 67.74\% at initial learning rate 0.01, and 69.83\% at initial learning rate 0.1. In the case of MobileNet-V2 on CIFAR-100 at low learning rate, we observe that the ``winning ticket'' can outperform the random reinitialization but failed to restore the baseline accuracy (73.10\%). This indicates the low learning rate is not desirable. For all illustrated cases, the ``winning ticket'''s accuracy is close to the random reinitialization at the initial learning rate 0.1. While in the case of learning rate 0.01, the ``winning ticket'' can outperform the random reinitialization over different sparsity ratios. Note there is a clearly accuracy gap between the pretrained DNNs with the initial learning rate 0.1 and with the initial learning rate 0.01.

From these experiments, the winning property exists at a low learning rate but does not exist at a relatively high learning rate. However, we would like to point out that the relatively high learning rate of 0.1 (which is, in fact, the standard learning rate on these datasets) results in \textit{notably higher accuracy} in the pretrained DNNs than the low learning rate (MobiletNet-V2 on CIFAR-10 93.86\% vs. 92.20\%, MobiletNet-V2 on CIFAR-100 74.76\% vs. 73.10\%, VGG-11 on CIFAR-100 69.83\% vs. 67.74\%, ResNet-20 on CIFAR-100 67.15\% vs. 63.10\%). We should not draw conclusion basd on the low (insufficient) learning rate in general.

\section{Weight Correlation in DNN Pre-Training}
\label{appendix:secB}


\begin{figure*}[t]
	\centering
	\begin{minipage}[b]{0.95\textwidth}
	\subfigure[VGG-11 for CIFAR-10]{
		\includegraphics[width=0.47\textwidth]{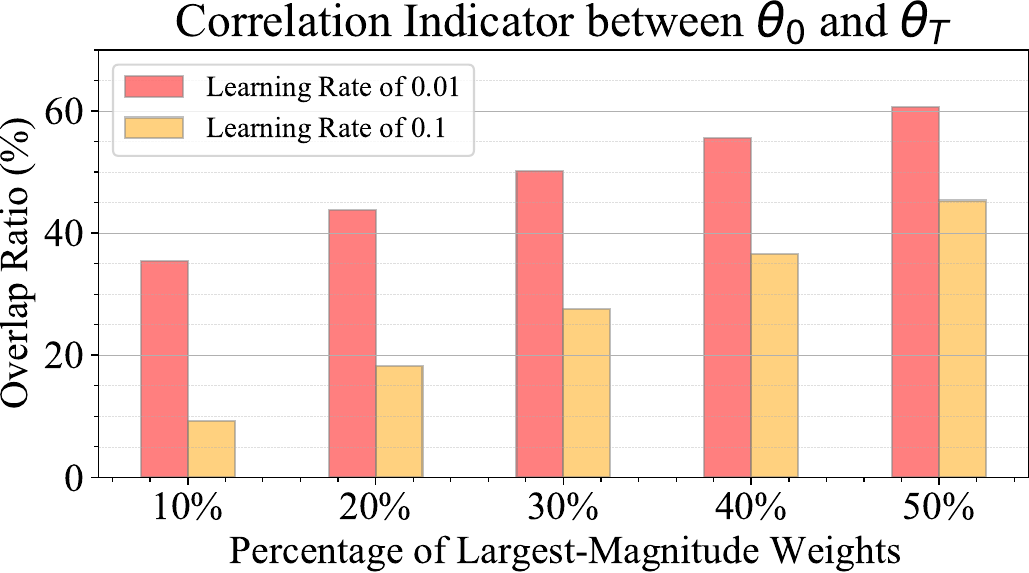}
		\label{appendixfig:fig4_vgg11_cifar10}
	}
	\hfill
	\subfigure[VGG-11 for CIFAR-100]{
		\includegraphics[width=0.47\textwidth]{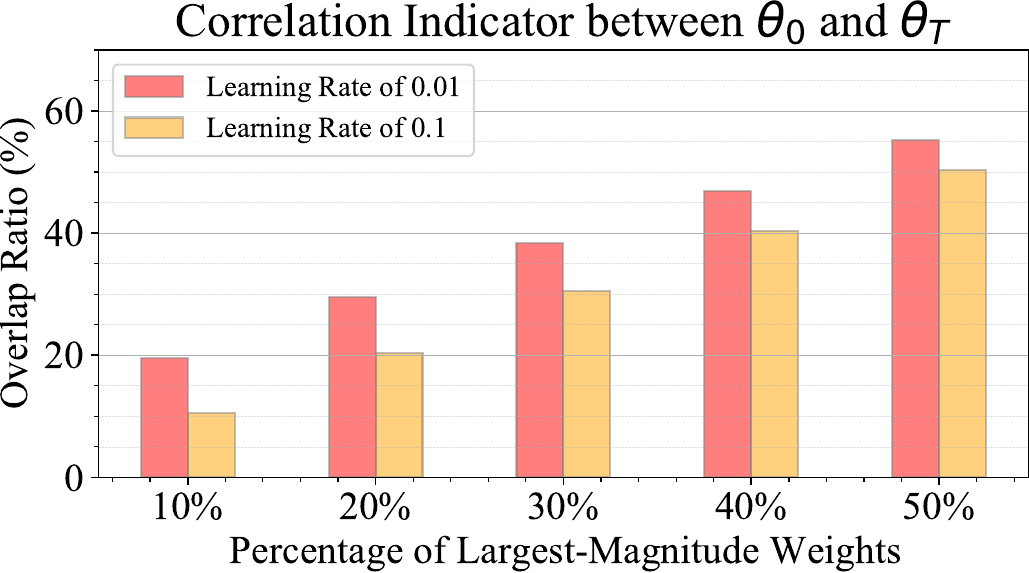}
		\label{appendixfig:fig4_vgg11_cifar100}
	}
	\subfigure[MobileNet-V2 for CIFAR-10]{
		\includegraphics[width=0.47\textwidth]{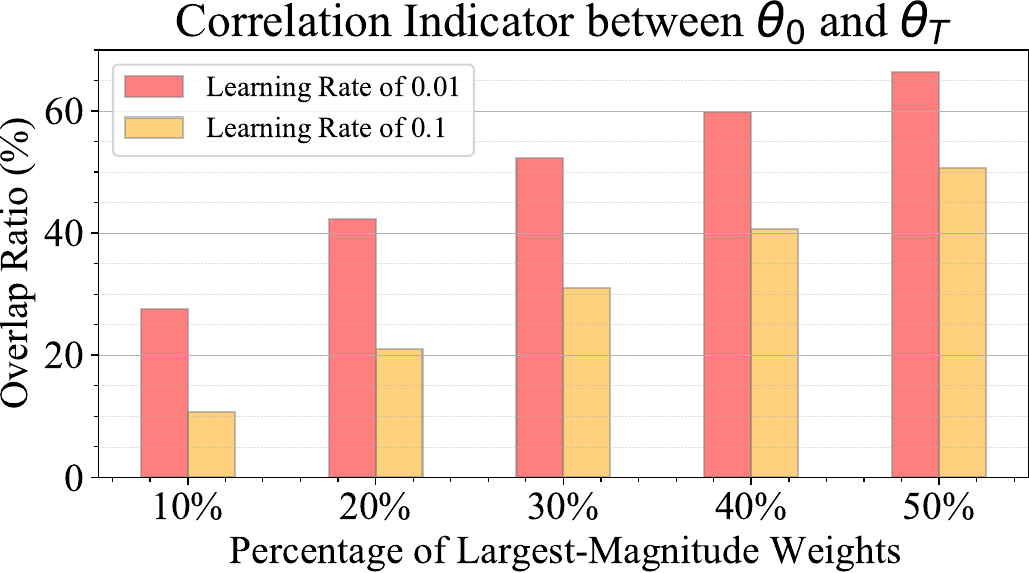}
		\label{appendixfig:fig4_mbnet_cifar10}
	}
	\hfill
	\subfigure[MobileNet-V2 for CIFAR-100]{
		\includegraphics[width=0.47\textwidth]{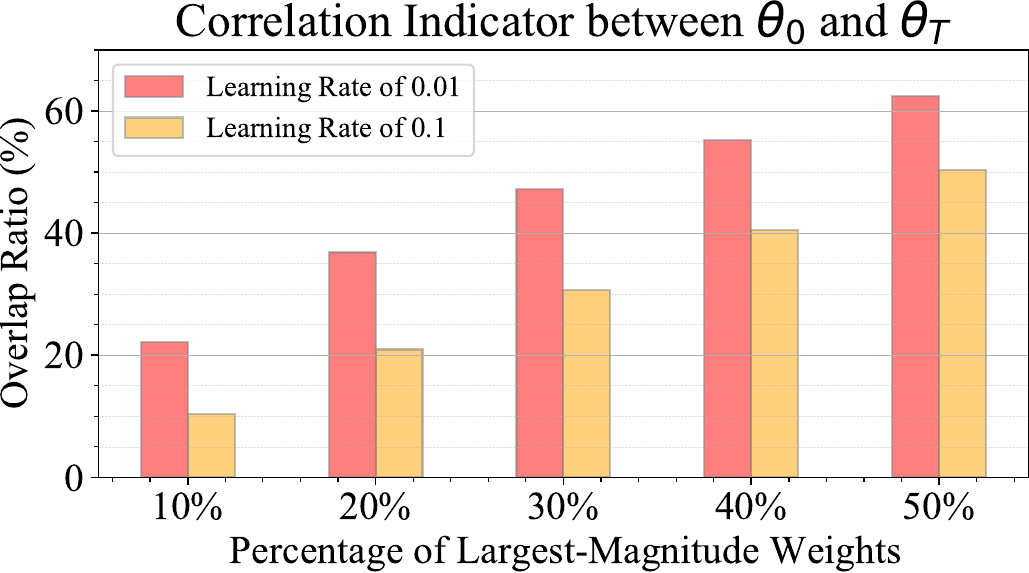}
		\label{appendixfig:fig4_mbnet_cifar100}
	}
	\end{minipage}
	\caption{The overlap ratios (when $p=$ 10\%, 20\%, 30\%, 40\% and 50\%) between the initial weights $\theta_0$ and the pretrained weights $\theta_T$ at learning rate of 0.01 and 0.1 on VGG-11 and MobileNet-V2 using CIFAR-10/100.}
	\label{appendixfig:fig4overlapscratch}
\end{figure*}

We investigate the \emph{correlation indicator} between the initial weights $\theta_0$ and the trained weights $\theta_T$ from DNN pretraining on VGG-11, ResNet-20, and MobileNet-V2 on CIFAR-10 and CIFAR-100 under learning rates of 0.01 and 0.1, respectively.


We have performed experiments to derive $R_p(\theta_0,\theta_T)$ on different DNN pretraining with different initial learning rates. Figure~\ref{appendixfig:fig4overlapscratch} illustrates the correlation indicator between the initial weights $\theta_0$ and the trained weights $\theta_T$ from DNN pretraining at learning rates of 0.01 and 0.1 on VGG-11 and MobileNet-V2 using CIFAR-10/100, respectively. 
We use the same hyperparameters mentioned in the setup without additional training tricks. Figure~\ref{appendixfig:fig4_vgg11_cifar10} and ~\ref{appendixfig:fig4_vgg11_cifar100} illustrate the result on VGG-11 for CIFAR-10/100. Figure~\ref{appendixfig:fig4_mbnet_cifar10} and ~\ref{appendixfig:fig4_mbnet_cifar100} illustrate the result on MobileNet-V2 for CIFAR-10/100. 

We can observe that $R_p(\theta_0,\theta_T)$ at a learning rate of 0.01 has a notably higher correlation compared to the case of learning rate 0.1. This observation indicates that the large-magnitude weights of $\theta_0$ are not fully updated at a low learning rate of 0.01, indicating that the pre-trained DNN is not well-trained. In the case of learning rate 0.1, the weights are sufficiently updated thus largely independent from the initial weights ($R_p(\theta_0,\theta_T)\approx p$, where $p=10\%, 20\%, 30\%, 40\%, 50\%$), indicating a well-trained DNN.


\section{Pruning \& Fine-tuning}
\label{appendix:secC}

\begin{figure}[h!]
	\centering
	\begin{minipage}[b]{0.95\columnwidth}
	\subfigure[MobileNet-V2 for CIFAR-10 with learning rate 0.01.]{
		\includegraphics[width=0.45\columnwidth]{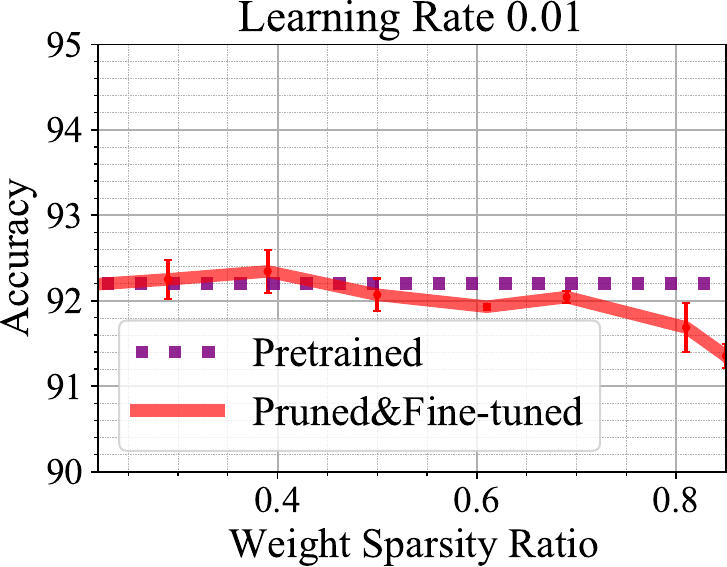}
		\label{appendixfig:fig5ab_mbnet_0.01_cifar10}
	}
	\hfill
	\subfigure[MobileNet-V2 for CIFAR-10 with learning rate 0.1.]{
		\includegraphics[width=0.45\columnwidth]{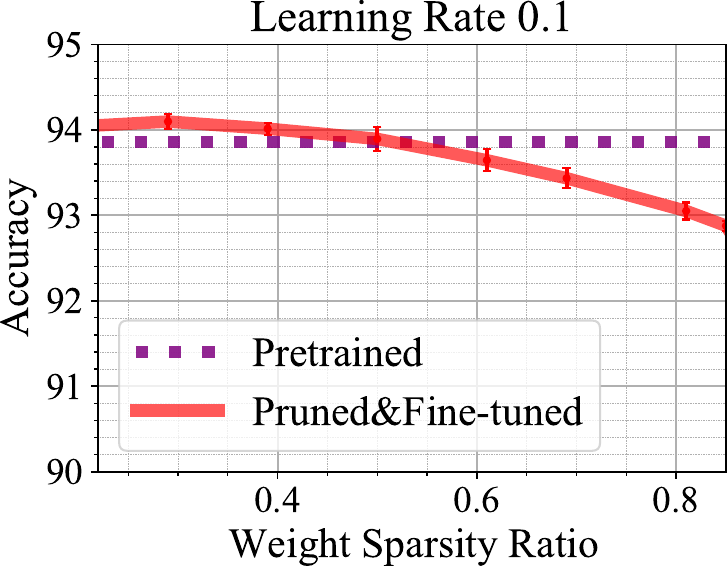}
		\label{appendixfig:fig5ab_mbnet_0.1_cifar10}
	}
	\subfigure[MobileNet-V2 for CIFAR-100 with learning rate 0.01.]{
		\includegraphics[width=0.45\columnwidth]{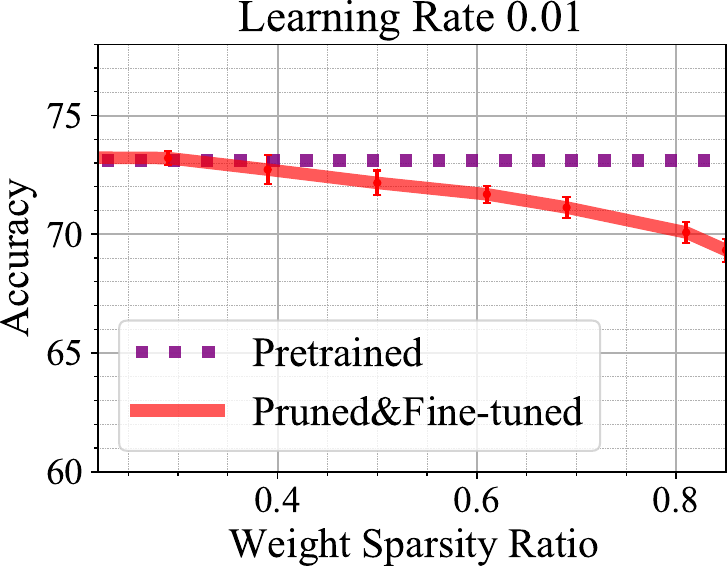}
		\label{appendixfig:fig5ab_mbnet_0.01_cifar100}
	}
	\hfill
    \subfigure[MobileNet-V2 for CIFAR-100 with learning rate 0.1.]{
		\includegraphics[width=0.45\columnwidth]{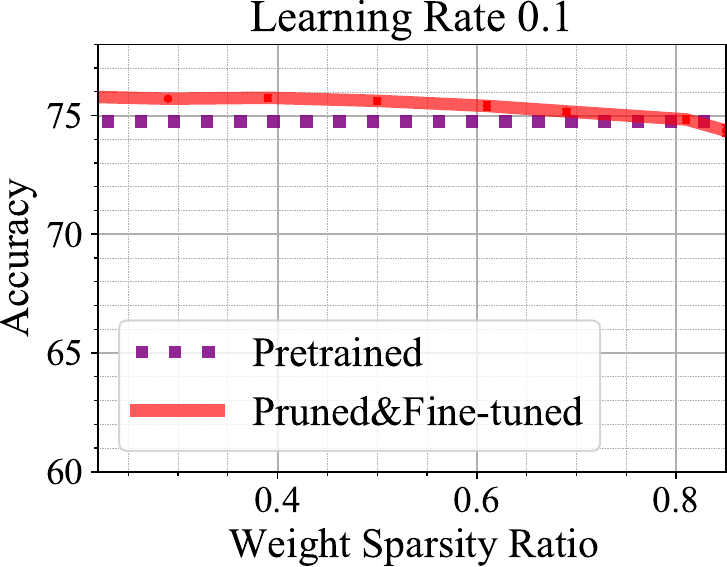}
		\label{appendixfig:fig5ab_mbnet_0.1_cifar100}
	}
	\end{minipage}
	\caption{Accuracy of $f\big(x;(\theta_{T}\odot m)_{T}\big)$ (``pruning \& fine-tuning") at different sparsity ratios with masks generated by iterative pruning on MobileNet-V2 using CIFAR-10/100.}
	\label{appendixfig:fig5ab_iterative_finetune}
\end{figure}

Consider the ``pruning \& fine-tuning" case formally defined in Section~\ref{sec:notation}, in which we apply mask $m$ on the trained weights $\theta_T$ from DNN pretraining, and then perform fine-tuning for another $T$ epochs. The final weights are denoted by $(\theta_{T}\odot m)_{T}$. We study accuracy of the ``pruning \& fine-tuning" result $f\big(x;(\theta_{T}\odot m)_{T}\big)$ at different sparsity ratios, with learning rates of 0.01 and 0.1 on different DNNs using CIFAR-10 and CIFAR-100. We use the same hyperparameters as mentioned in the setup ($T=150$). The accuracies of the pretrained DNNs with corresponding learning rates are also provided. 
Figure~\ref{appendixfig:fig5ab_mbnet_0.01_cifar10} and~\ref{appendixfig:fig5ab_mbnet_0.1_cifar10} illustrate the ``pruning \& fine-tuning" result on MobileNet-V2 for CIFAR-10 using learning rates of 0.01 and 0.1, respectively.
Figure~\ref{appendixfig:fig5ab_mbnet_0.01_cifar100} and~\ref{appendixfig:fig5ab_mbnet_0.1_cifar100} illustrate the ``pruning \& fine-tuning" result on MobileNet-V2 for CIFAR-100 with learning rates of 0.01 and 0.1, respectively. In the case of MobileNet-V2 for CIFAR-100 with the initial learning rate 0.1, the ``pruning \& fine-tuning" scheme consistently perform better than the pretrained dense DNN (74.76\%).

We can observe that $f\big(x;(\theta_{T}\odot m)_{T}\big)$ achieves relatively high accuracy, close to or higher than the accuracy of the pretrained DNN at the same learning rate (even at the desirable learning rate 0.1).

\section{Sparse Correlation}
\label{appendix:secD}

\begin{figure}[!b]
	\centering
	\begin{minipage}[b]{1\columnwidth}
	\subfigure[ResNet-20 for CIFAR-100 with learning rate 0.01.]{
		\includegraphics[width=0.45\columnwidth]{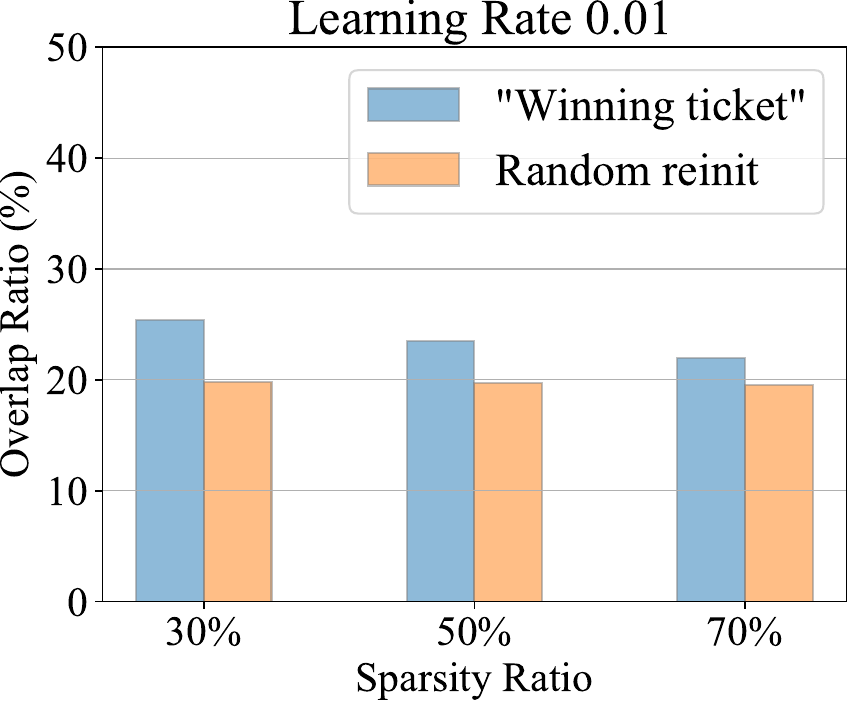}
		\label{appendixfig:fig5cd_resnet20_cifar100_0.01}
	}
	\hfill
	\subfigure[ResNet-20 for CIFAR-100 with learning rate 0.1.]{
		\includegraphics[width=0.45\columnwidth]{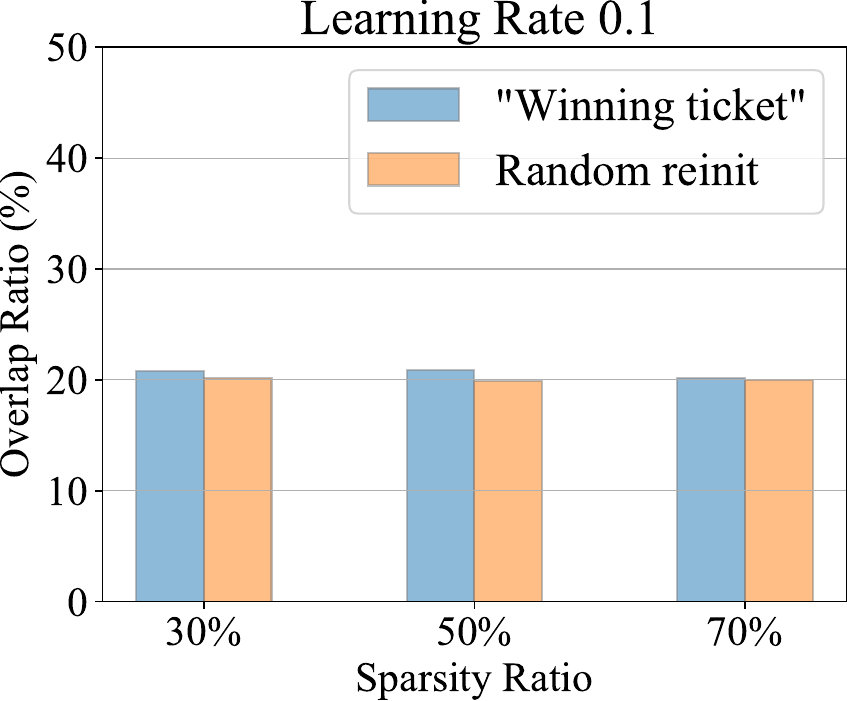}
		\label{appendixfig:fig5cd_resnet20_cifar100_0.1}
	}
	\end{minipage}
	\caption{The weight correlation (overlap ratio) comparison at $p=0.2$, between $\theta_0\odot m$ (``winning ticket'') and $(\theta_T\odot m)_T$ (pruned\&fine-tuned weights), and between $\theta_0^{\prime}\odot m$ (random reinitialization) and $(\theta_T\odot m)_T$ (pruned\&fine-tuned weights) under 0.3, 0.5, 0.7 sparsity ratios on ResNet-20 using CIFAR-100.}
	\label{appendixfig:fig5cd_resnet20}
\end{figure}

\begin{figure}[t!]
	\centering
	\begin{minipage}[b]{1\columnwidth}
	\subfigure[VGG-11 for CIFAR-10 with learning rate 0.01.]{
		\includegraphics[width=0.45\columnwidth]{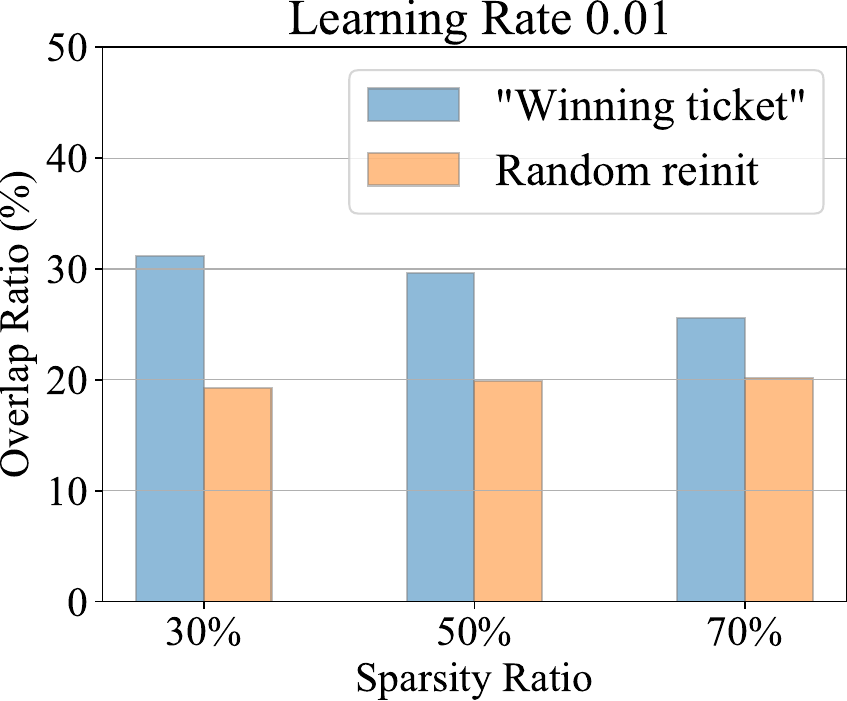}
		\label{appendixfig:fig5cd_vgg11_cifar10_0.01}
	}
	\hfill
	\subfigure[VGG-11 for CIFAR-10 with learning rate 0.1.]{
		\includegraphics[width=0.45\columnwidth]{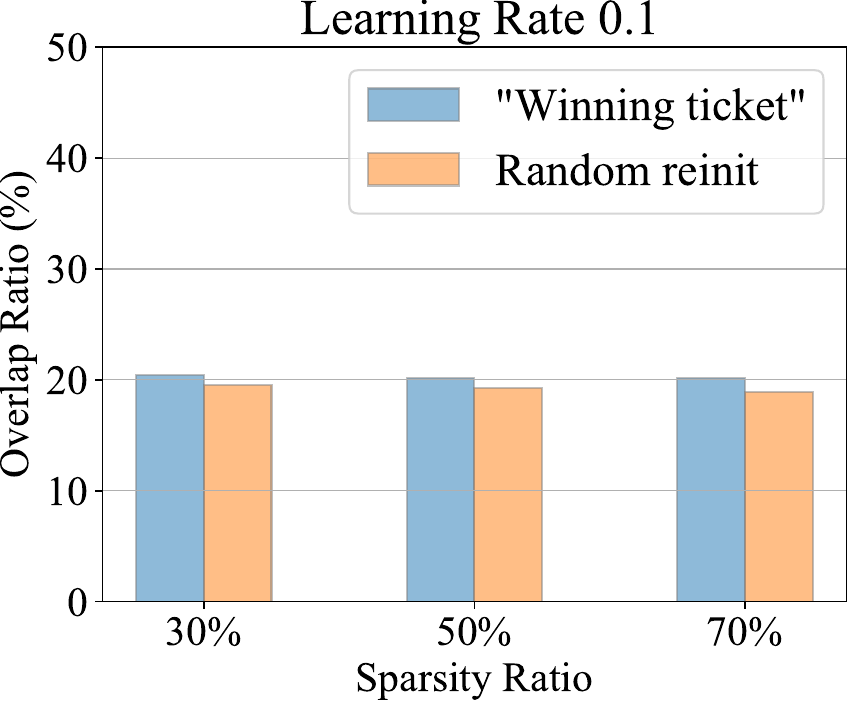}
		\label{appendixfig:fig5cd_vgg11_cifar10_0.1}
	}
	\subfigure[VGG-11 for CIFAR-100 with learning rate 0.01.]{
		\includegraphics[width=0.45\columnwidth]{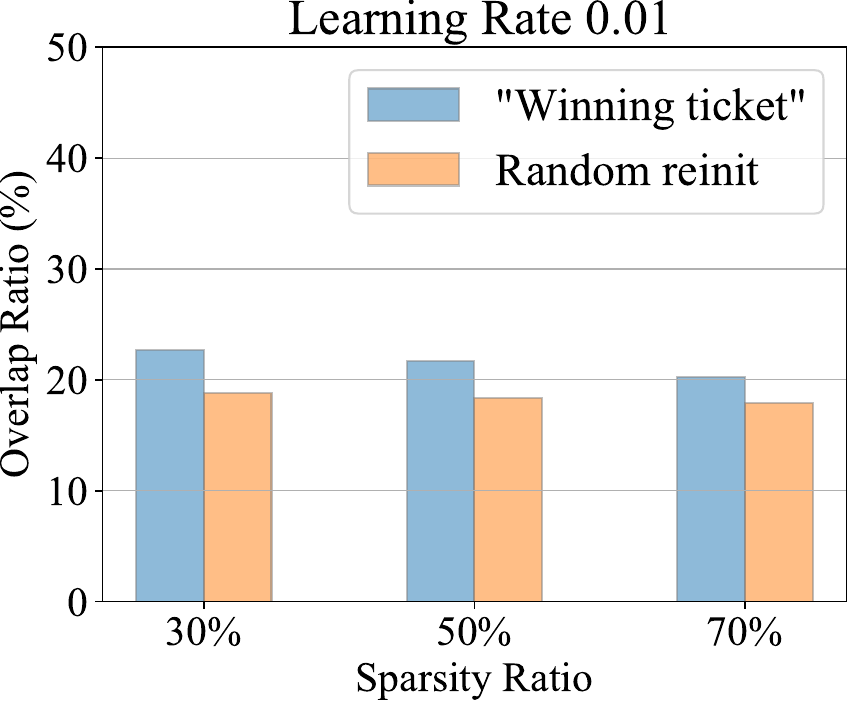}
		\label{appendixfig:fig5cd_vgg11_cifar100_0.01}
	}
	\hfill
    \subfigure[VGG-11 for CIFAR-100 with learning rate 0.1.]{
		\includegraphics[width=0.45\columnwidth]{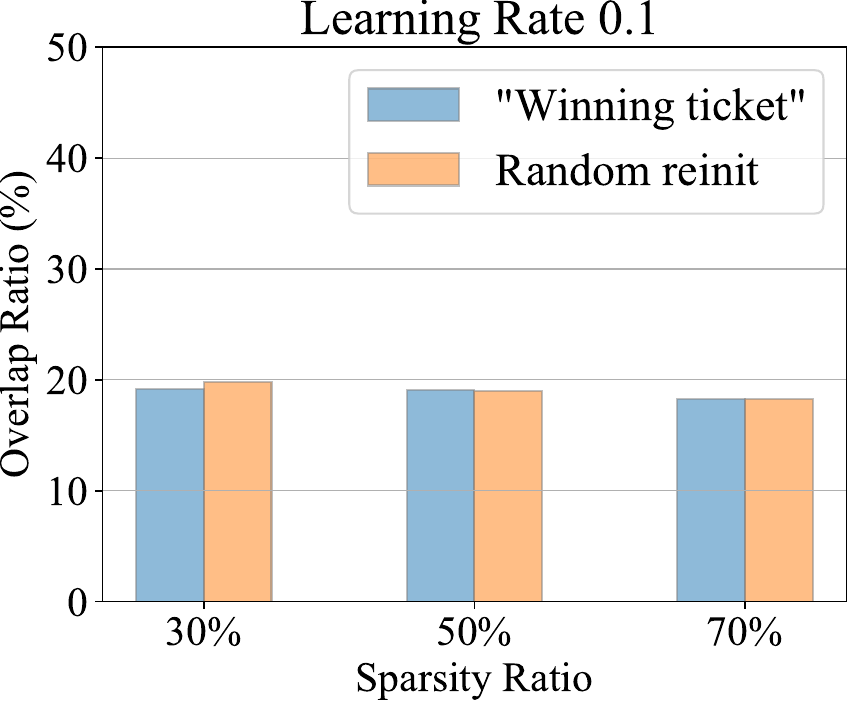}
		\label{appendixfig:fig5cd_vgg11_cifar100_0.1}
	}
	\end{minipage}
	\caption{The weight correlation (overlap ratio) comparison at $p=0.2$, between $\theta_0\odot m$ (``winning ticket'') and $(\theta_T\odot m)_T$ (pruned\&fine-tuned weights), and between $\theta_0^{\prime}\odot m$ (random reinitialization) and $(\theta_T\odot m)_T$ (pruned\&fine-tuned weights) under 0.3, 0.5, 0.7 sparsity ratios on VGG-11 using CIFAR-10/100.}
	\label{appendixfig:fig5cd_vgg11}
\end{figure}

\begin{figure}[t!]
	\centering
	\begin{minipage}[b]{1\columnwidth}
	\subfigure[MobileNet-V2 for CIFAR-10 with learning rate 0.01.]{
		\includegraphics[width=0.45\columnwidth]{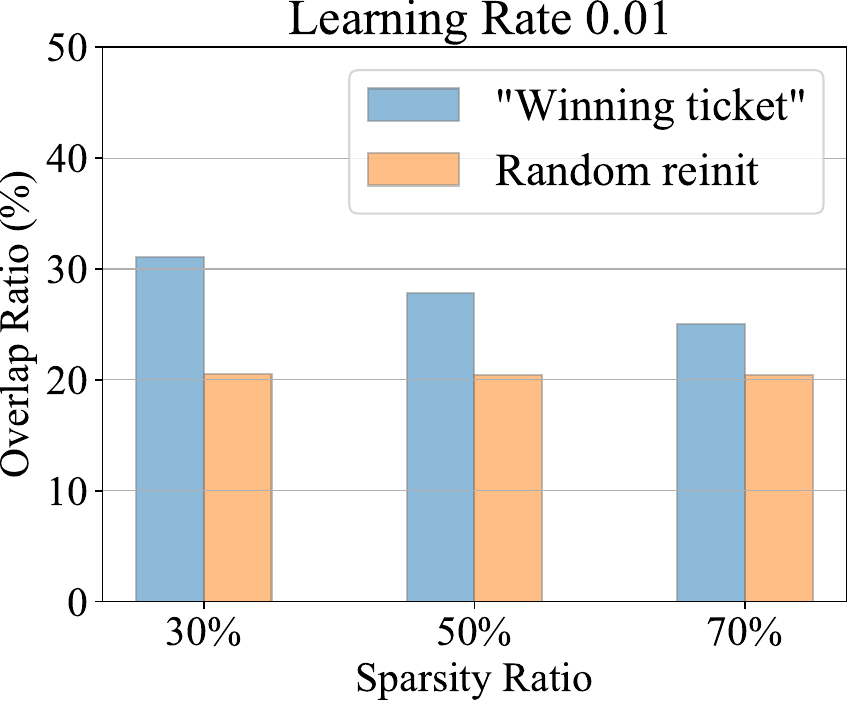}
		\label{appendixfig:fig5cd_mbnet_cifar10_0.01}
	}
	\hfill
	\subfigure[MobileNet-V2 for CIFAR-10 with learning rate 0.1.]{
		\includegraphics[width=0.45\columnwidth]{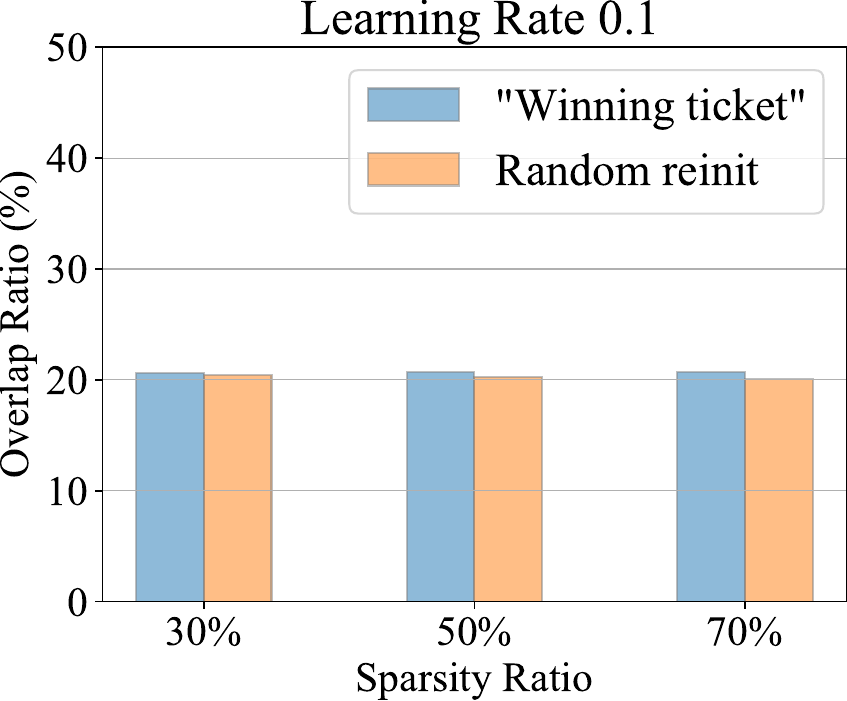}
		\label{appendixfig:fig5cd_mbnet_cifar10_0.1}
	}
	\subfigure[MobileNet-V2 for CIFAR-100 with learning rate 0.01.]{
		\includegraphics[width=0.45\columnwidth]{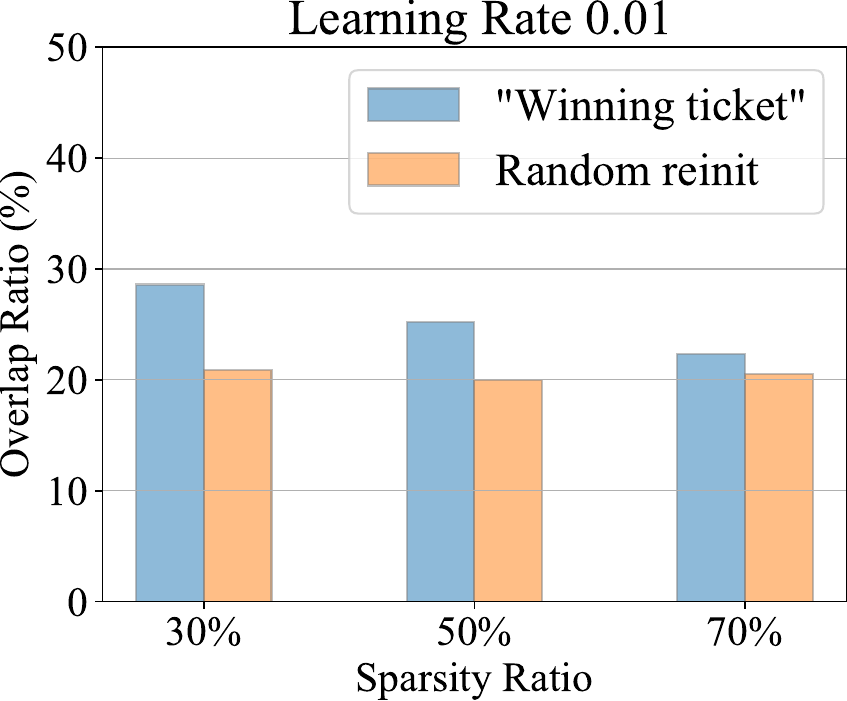}
		\label{appendixfig:fig5cd_mbnet_cifar100_0.01}
	}
	\hfill
    \subfigure[MobileNet-V2 for CIFAR-100 with learning rate 0.1.]{
		\includegraphics[width=0.45\columnwidth]{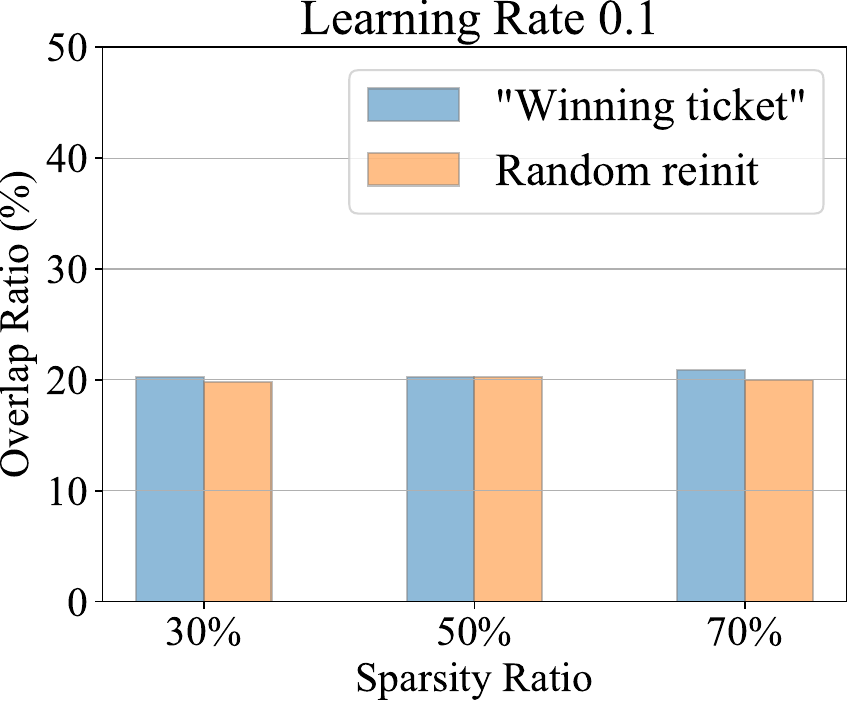}
		\label{appendixfig:fig5cd_mbnet_cifar100_0.1}
	}
	\end{minipage}
	\caption{The weight correlation (overlap ratio) comparison at $p=0.2$, between $\theta_0\odot m$ (``winning ticket'') and $(\theta_T\odot m)_T$ (pruned\&fine-tuned weights), and between $\theta_0^{\prime}\odot m$ (random reinitialization) and $(\theta_T\odot m)_T$ (pruned\&fine-tuned weights) under 0.3, 0.5, 0.7 sparsity ratios on MobileNet-V2 using CIFAR-10/100.}
	\label{appendixfig:fig5cd_mbnet}
\end{figure}

We study the correlation between $\theta_{0}\odot m$ ($\theta'_{0}\odot m$) and $(\theta_{T}\odot m)_{T}$ to shed some light on the cause of winning property. We illustrate the correlation on ResNet-20, VGG-11 and MobileNet-V2 for CIFAR-10/100 at learning rate 0.01, 0.1, respectively. We show the correlation indicator between $\theta_{0}\odot m$ (``winning ticket'') and $(\theta_{T}\odot m)_{T}$, and between $\theta'_{0}\odot m$ (random reinitialization) and $(\theta_{T}\odot m)_{T}$ at learning rate 0.01, 0.1. Figure~\ref{appendixfig:fig5cd_resnet20} illustrates the result of ResNet-20 for CIFAR-100 at the learning rate 0.01 and 0.1. Figure~\ref{appendixfig:fig5cd_vgg11} shows the result of VGG-11 for CIFAR-10/100 and Figure~\ref{appendixfig:fig5cd_mbnet} shows the result of MobileNet-V2 for CIFAR-10/100 at learning rates 0.01, 0.1, respectively. In the case of high learning rate 0.1, the weight correlation between $\theta_0\odot m$ (``winning ticket'') and $(\theta_T\odot m)_T$ (pruned\&fine-tuned weights), and between $\theta_0^{\prime}\odot m$ (random reinitialization) and $(\theta_T\odot m)_T$ (pruned\&fine-tuned weights) are similar (and minor) under different sparsity ratios.

From these results we can observe the positive correlation between $\theta_{0}\odot m$ and $(\theta_{T}\odot m)_{T}$ at the low learning rate, when the winning property exists. Such correlation is minor in the other cases.


\begin{figure*}[!h]
    \centering
    \begin{subfigure}
        \centering
        {\includegraphics[width=0.32\textwidth]{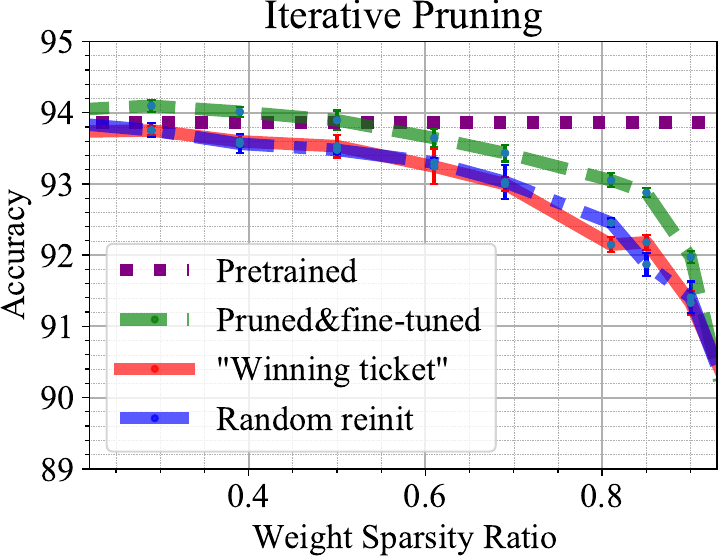}
        \includegraphics[width=0.32\textwidth]{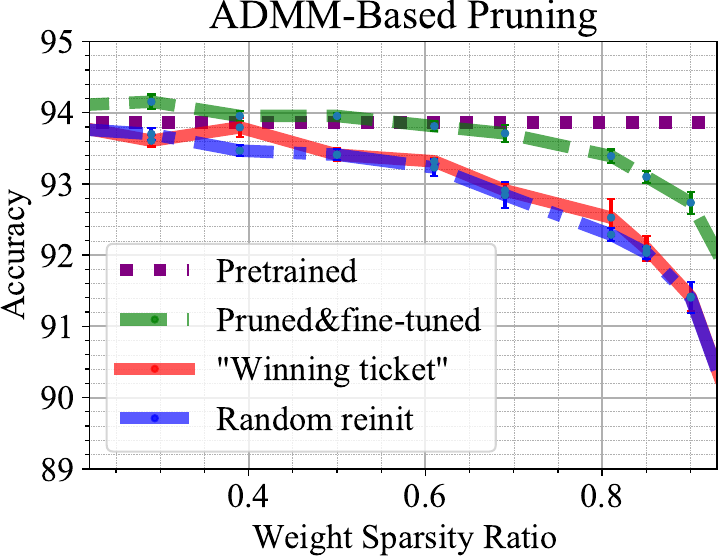}
        \includegraphics[width=0.32\textwidth]{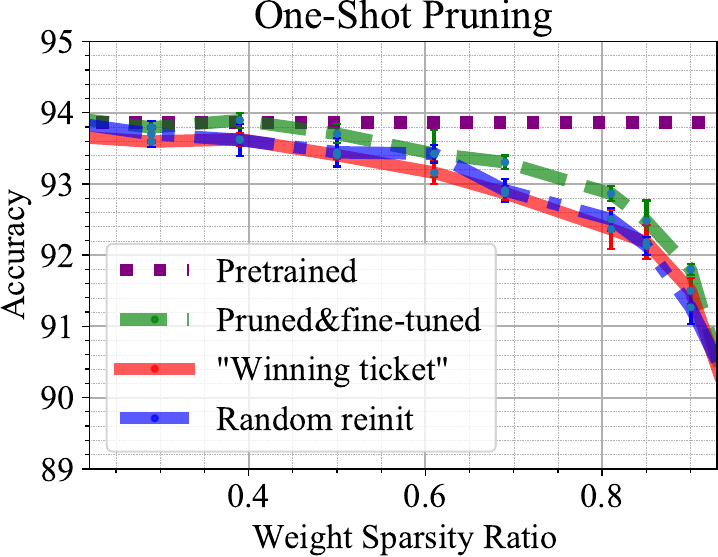}}
        \caption{Accuracy of pruning \& fine-tuning vs. two sparse training cases (``winning ticket'' and random reinitialization) on MobileNet-V2 using CIFAR-10.}\label{appendixfig:fig6_mbnet_cifar10}
    \end{subfigure}
    \vskip\baselineskip
    \begin{subfigure}
        \centering
        \includegraphics[width=0.32\textwidth]{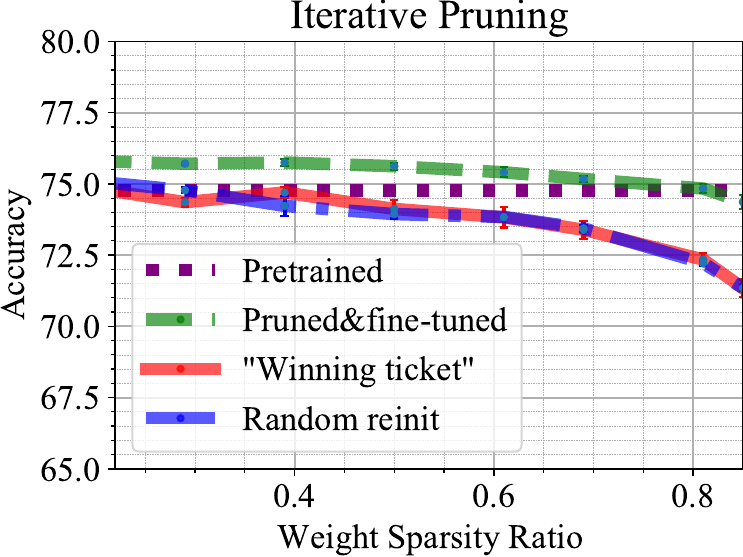}
        \includegraphics[width=0.32\textwidth]{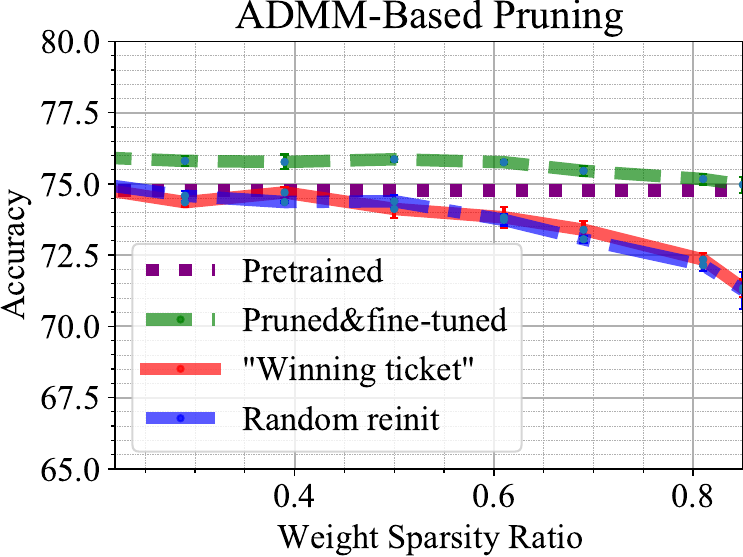}
        \includegraphics[width=0.32\textwidth]{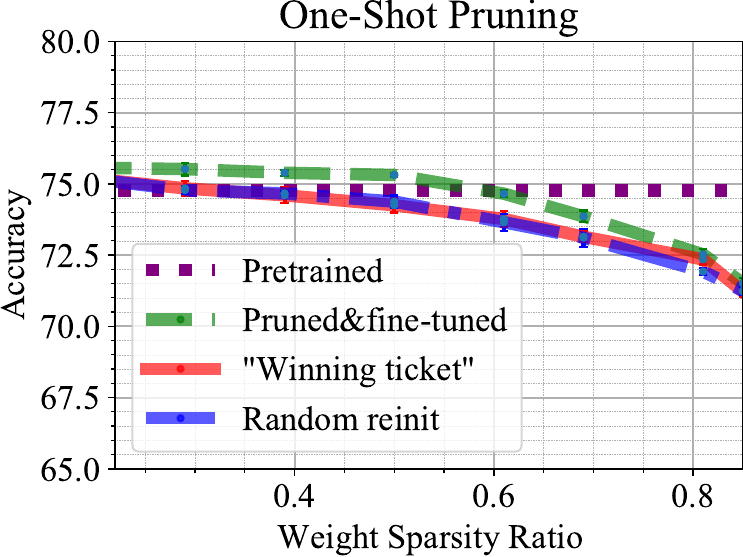}
        \caption{Accuracy of pruning \& fine-tuning vs. two sparse training cases (``winning ticket'' and random reinitialization) on MobileNet-V2 using CIFAR-100.}\label{appendixfig:fig6_mbnet_cifar100}
    \end{subfigure}
    \vskip\baselineskip
    \begin{subfigure}
        \centering
        \includegraphics[width=0.32\textwidth]{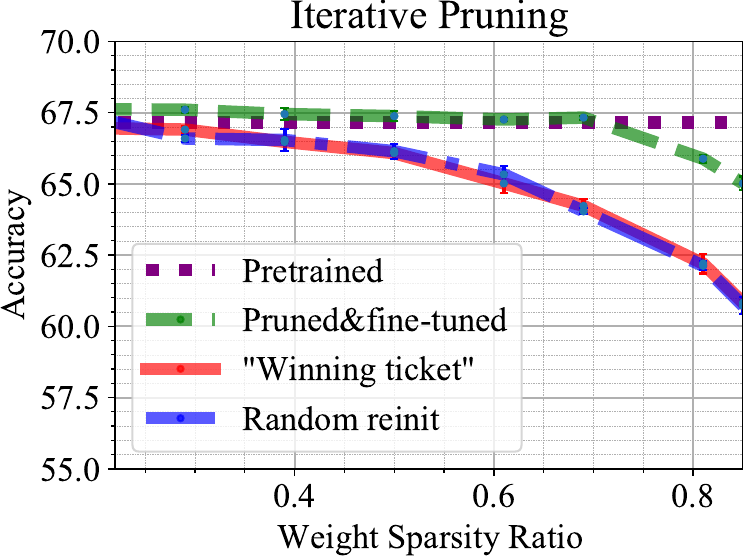}
        \includegraphics[width=0.32\textwidth]{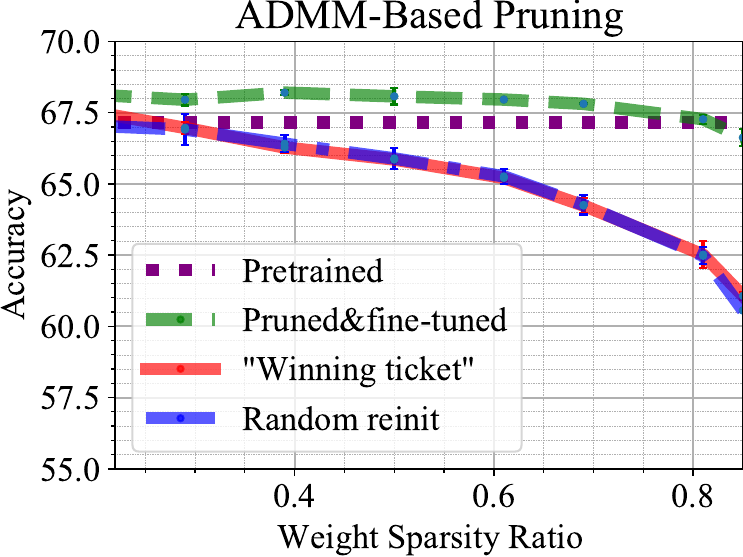}
        \includegraphics[width=0.32\textwidth]{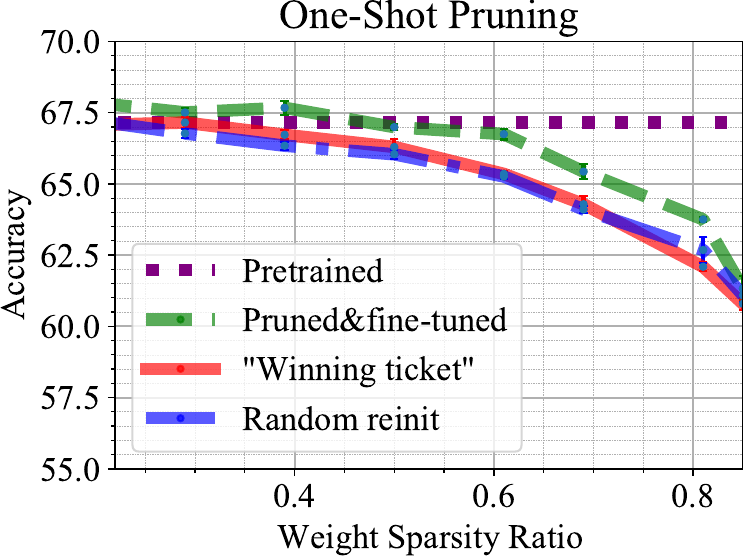}
        \caption{Accuracy of pruning \& fine-tuning vs. two sparse training cases (``winning ticket'' and random reinitialization) on ResNet-20 using CIFAR-100.}\label{appendixfig:fig6_resnet20_cifar100}
    \end{subfigure} 
    \vskip\baselineskip
    \begin{subfigure}
        \centering
        \includegraphics[width=0.32\textwidth]{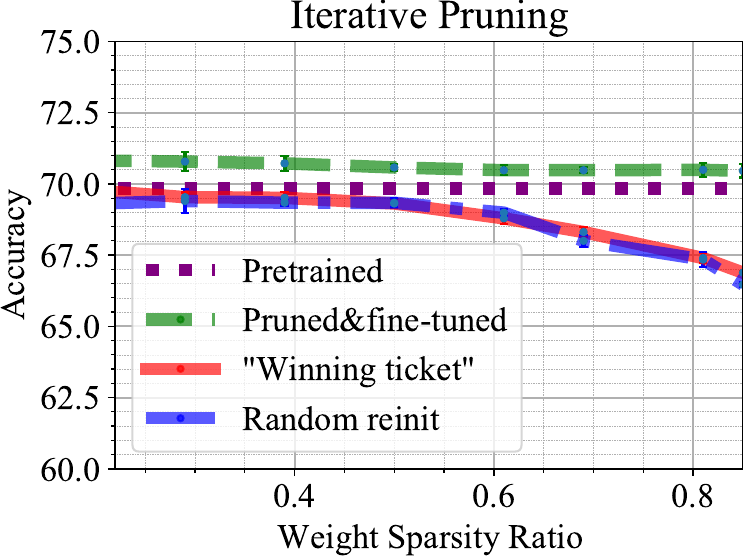}
        \includegraphics[width=0.32\textwidth]{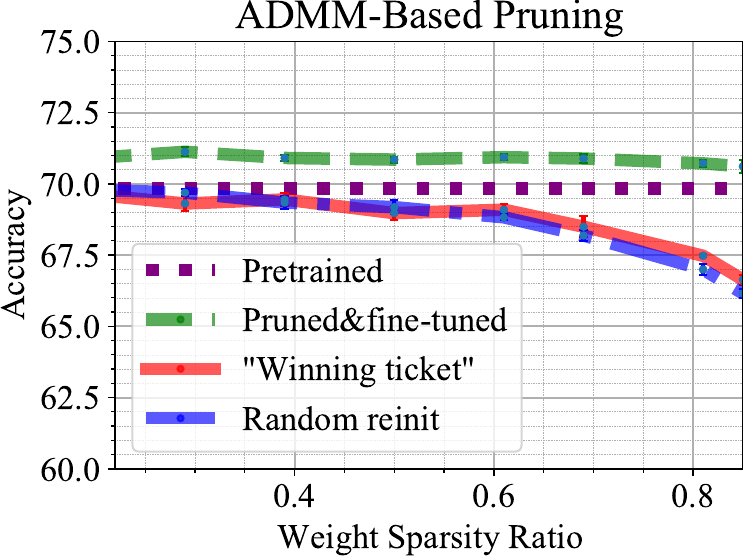}
        \includegraphics[width=0.32\textwidth]{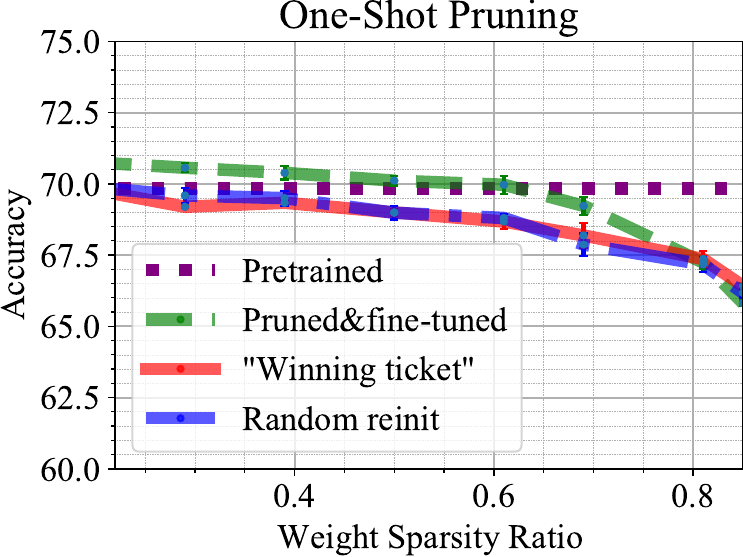}
        \caption{Accuracy of pruning \& fine-tuning vs. two sparse training cases (``winning ticket'' and random reinitialization) on VGG-11 using CIFAR-100.}\label{appendixfig:fig6_vgg11_cifar100}
    \end{subfigure}
\end{figure*}

\begin{figure*}[!h]
	\centering
	\begin{minipage}[b]{0.97\textwidth}
	\subfigure[MobileNet-V2 on CIFAR-10 at learning rate 0.1.]{
		\includegraphics[width=0.47\textwidth]{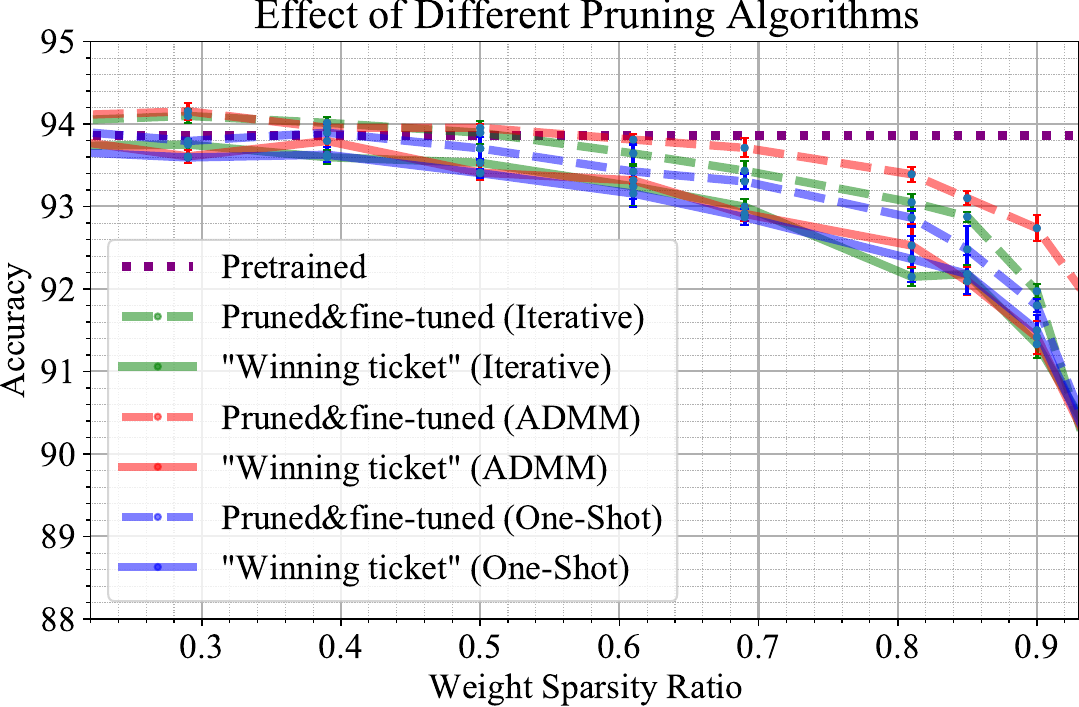}
		\label{appendixfig:fig7_mbnet_cifar10}
	}
	\hfill
	\subfigure[MobileNet-V2 on CIFAR-100 at learning rate 0.1.]{
		\includegraphics[width=0.47\textwidth]{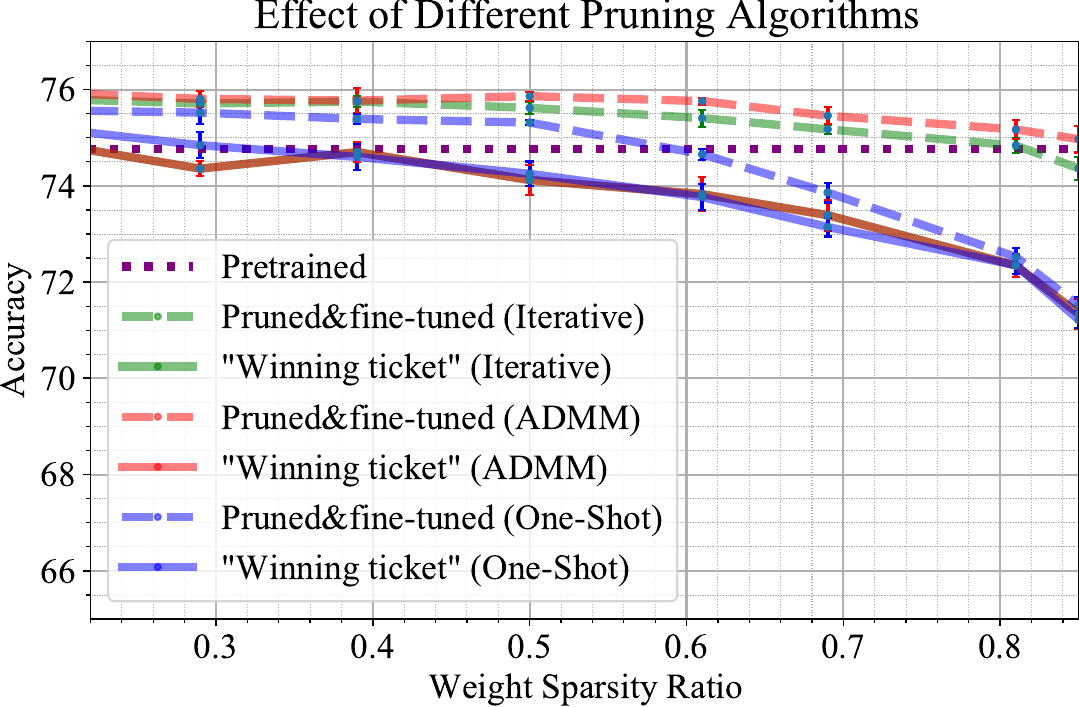}
		\label{appendixfig:fig7_mbnet_cifar100}
	}
	\end{minipage}
	\caption{Accuracy of pruning \& fine-tuning and sparse training (``winning ticket'' case), under all three pruning algorithms (iterative pruning, ADMM-based pruning, and one-shot pruning) for mask generation on MobileNet-V2 using CIFAR-10/100.}
	\label{appendixfig:fig7_prunefinetune_vs_ticket_mbnet_cifar10_100}
\end{figure*}

\begin{figure*}[!h]
	\centering
	\begin{minipage}[b]{0.97\textwidth}
	\subfigure[ResNet-20 on CIFAR-100 at learning rate 0.1.]{
		\includegraphics[width=0.47\textwidth]{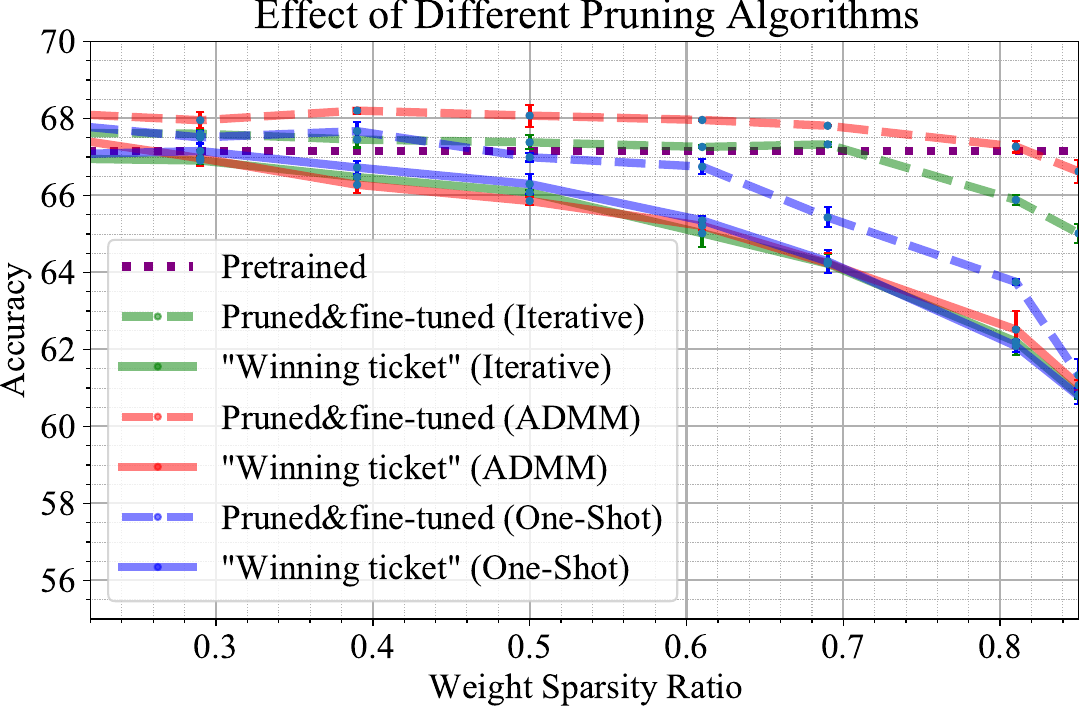}
		\label{appendixfig:fig7_resnet_cifar100}
	}
	\hfill
	\subfigure[VGG-11 on CIFAR-100 at learning rate 0.1.]{
		\includegraphics[width=0.47\textwidth]{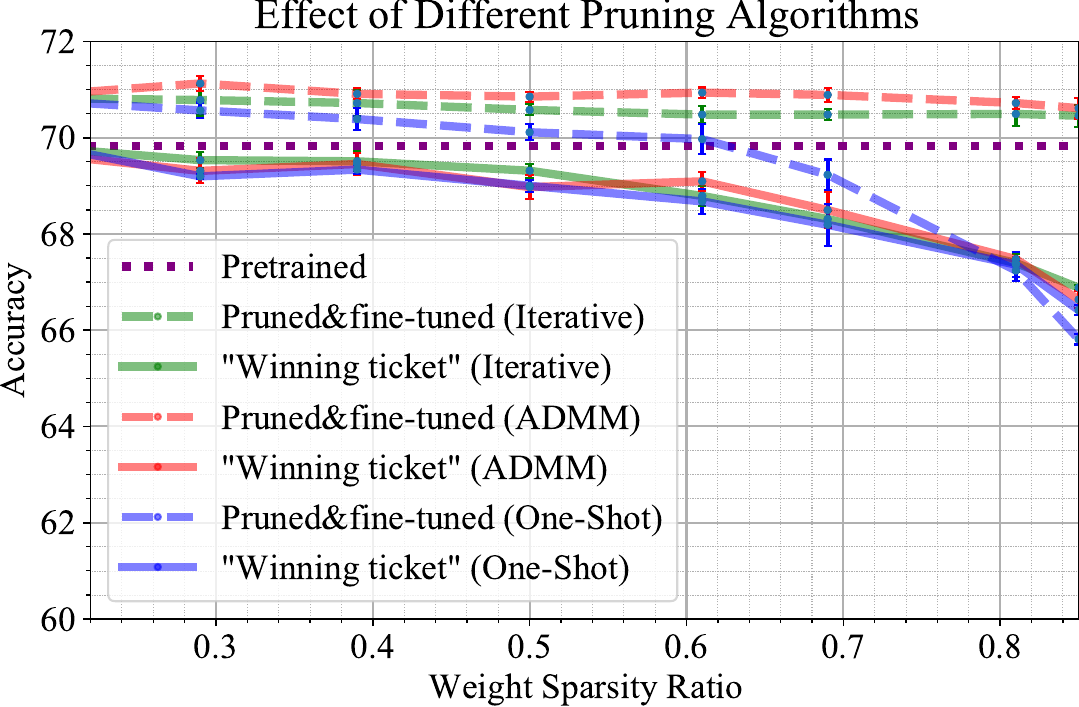}
		\label{appendixfig:fig7_vgg_cifar100}
	}
	\end{minipage}
	\caption{Accuracy of pruning \& fine-tuning and sparse training (``winning ticket'' case), under all three pruning algorithms (iterative pruning, ADMM-based pruning, and one-shot pruning) for mask generation on ResNet-20 and VGG-11 using CIFAR-100.}
	\label{appendixfig:fig7_prunefinetune_vs_ticket_res_vgg_cifar100}
\end{figure*}


\section{Different Pruning Algorithms}
\label{appendix:secE}

We explore the different pruning algorithms on ResNet-20, MobileNet-V2 and VGG-11 using CIFAR-10/100. We use the desirable learning rate 0.1, $T=150$ epochs, and the same hyperparameters introduced in Section~\ref{subsec:revisitLT}. 
We compare accuracy between pruning \& fine-tuning (i.e., training (fine-tuning) from $\theta_{T}\odot m$) and the two sparse training scenarios ``winning ticket'' (i.e., training from $\theta_{0}\odot m$) and random reinitialization (i.e., training from $\theta'_{0}\odot m$) at different sparsity ratios. We investigate three pruning algorithms to derive mask $m$: Iterative pruning algorithm, ADMM-based pruning \cite{zhang2018systematic} and one-shot pruning algorithm. We explore accuracy comparison results between pruning \& fine-tuning and the two sparsity training scenarios. Figure~\ref{appendixfig:fig6_mbnet_cifar10} and~\ref{appendixfig:fig6_mbnet_cifar100} illustrate the accuracy comparison on MobileNet-V2 using CIFAR-10 and CIFAR-100, respectively. 
Figure~\ref{appendixfig:fig6_resnet20_cifar100} shows the result on ResNet-20 using CIFAR-100.
Figure~\ref{appendixfig:fig6_vgg11_cifar100} illustrates the result on VGG-11 for CIFAR-100.

From these results, we can clearly observe the accuracy gap between pruning \& fine-tuning and the two sparse training cases (lottery ticket setting). For MobiletNet-V2 on CIFAR-100, with the masks generated from iterative pruning and ADMM-based pruning, the pruning \& fine-tuning scheme can consistently outperform the pretrained dense DNN up to sparsity ratio 85\%. Similarly results can be observed on VGG-11 using CIFAR-100. Meanwhile, at sparsity ratio 0.39 (39\%), the pruning \& fine-tuning scheme with mask generated from ADMM-based pruning can achieve accuracy 76.04\% while the pretrained DNN's accuracy is only 74.76\% (under the desirable learning rate 0.1).

We observe the notable advantage of pruning \& fine-tuning over the lottery ticket setting, even with a weak one-shot pruning algorithm for mask generation. Note there is no accuracy difference between the two sparse training cases. 
Pruning \& fine-tuning under ADMM-based pruning can restore the accuracy of pretrained DNN with the highest sparsity ratio compared to the other two pruning algorithms. 
Clearly, the consistent advantage of pruning \& fine-tuning is attributed to the fact that mask $m$ is applied to pretrained weights $\theta_T$ instead of the initialized weights $\theta_0$. In fact, information in $\theta_T$ is important for the sparse subnetwork to maintain accuracy of the pretrained dense network. Or in other words, weights in the desirable sparse subnetwork should have correlation with $\theta_T$ instead of $\theta_0$.

Further we evaluate the relative performance (accuracy) of these three pruning algorithms.
We combine the above results and demonstrate the accuracy performances of pruning \& fine-tuning and sparse training (``winning ticket'' case), under all three pruning algorithms. Figure~\ref{appendixfig:fig7_mbnet_cifar10} and~\ref{appendixfig:fig7_mbnet_cifar100} show the overall accuracy performance comparison on MobileNet-V2 using CIFAR-10 and CIFAR-100, respectively. Figure~\ref{appendixfig:fig7_resnet_cifar100} shows the result on ResNet-20 using CIFAR-100 and Figure~\ref{appendixfig:fig7_vgg_cifar100} shows the result on VGG-11 using CIFAR-100.

We observe the order in the accuracy performance: ADMM-based pruning on top, iterative pruning in the middle, and one-shot pruning the lowest. This order is the same for pruning \& fine-tuning and sparse training. Note that the pruning algorithm is utilized to generate mask $m$, while the other conditions are the same (i.e., $\theta_T$, fine-tuning $T$ epochs on $\theta_T\odot m$, or sparse training on $\theta_0\odot m$). Hence, the relative performance is attributed to the quality in mask generation. We can conclude that the selection of pruning algorithm is critical in generating the sparse subnetwork as the quality of mask generation plays a key role in the context of pruning scenario.

\begin{figure*}[!h]
	\centering
	\begin{minipage}[b]{1.0\textwidth}
	\subfigure[MobileNet-V2 for CIFAR-100 at learning rate of 0.1.]{
		\includegraphics[width=0.32\textwidth]{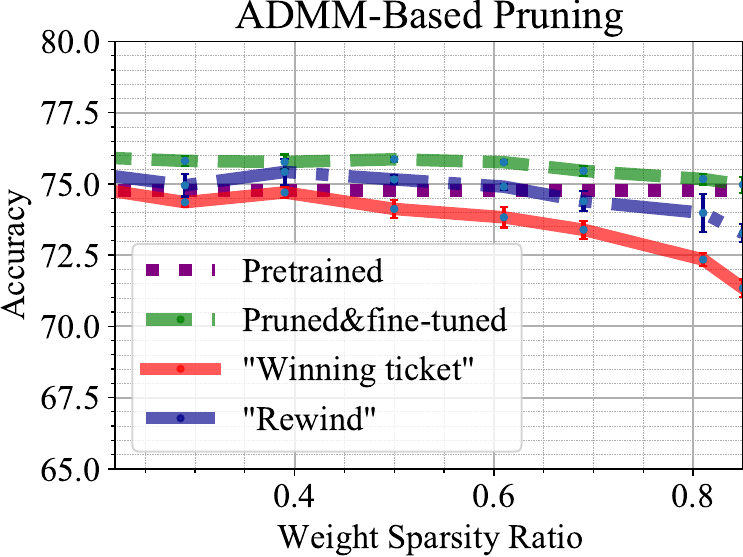}
		\label{appendixfig:fig8_rewind_mbnet_cifar100}
	}
	\subfigure[ResNet-20 for CIFAR-100 at learning rate of 0.1.]{
		\includegraphics[width=0.32\textwidth]{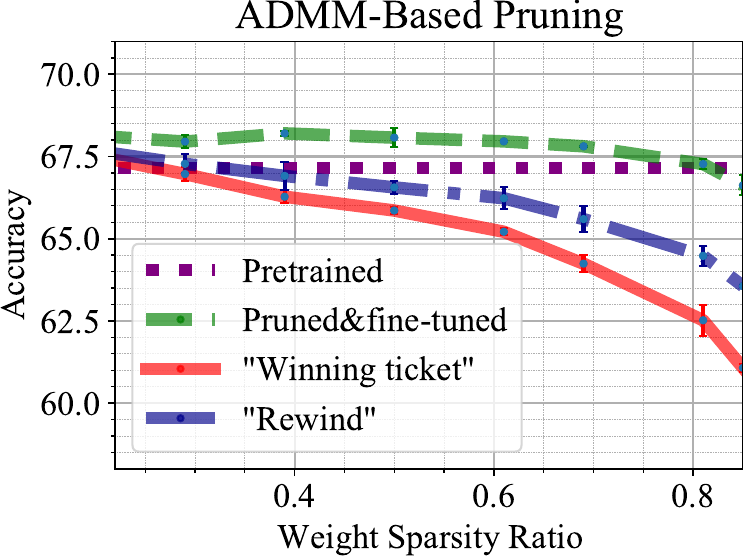}
		\label{appendixfig:fig8_rewind_resnet20_cifar100}
	}
	\subfigure[VGG-11 for CIFAR-100 at learning rate of 0.1.]{
		\includegraphics[width=0.32\textwidth]{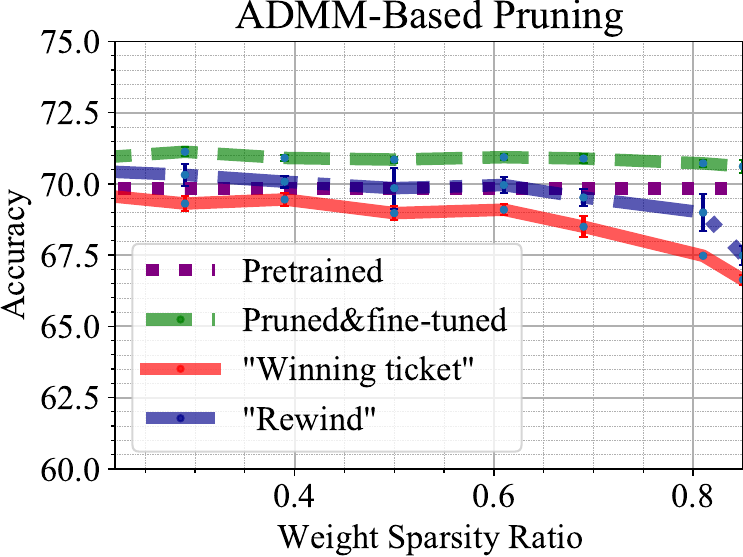}
		\label{appendixfig:fig8_rewind_vgg11_cifar100}
	}
	\end{minipage}
	\caption{Accuracy performance of $(\theta_T \odot m)_T$ (pruned\&fine-tuned), $(\theta_k \odot m)_T$ (``rewind'') and $(\theta_0 \odot m)_T$ (``winning ticket'') on MobileNet-V2, ResNet-20 and VGG-11 for CIFAR-100 over a range of different sparsity ratios. The masks are generated by ADMM-based pruning algorithm and the initial learning rate is 0.1. }
	\label{appendixfig:fig8_rewind}
\end{figure*}

\section{An Analysis from Weight Correlation Perspective}
\label{appendix:secF}

We provide the correlation between $(\theta_{T}\odot m)_{T}$ and $\theta_0$, and between $(\theta_{T}\odot m)_{T}$ and $\theta_T$ for VGG-11, ResNet-20 and MobileNet-V2 using CIFAR-10/100. The training epoch $T=150$ and the initial learning rate is 0.1. The masks are generated by ADMM-based pruning algorithm. Note that $\theta_0$ and $\theta_T$ are dense models, while $(\theta_{T}\odot m)_{T}$ is a sparse model. To utilize the \emph{correlation indicator}, we extend the correlation scenario of dense DNNs vs. dense DNNs to sparse DNNs vs. dense DNNs by restricting $p$ less than $(1-$sparsity ratio$)$ of sparse DNNs. In this experiment, we consider weight correlation at $p=0.2$ and the sparsity ratio is 0.50 (50\%) for the DNNs. The results are illustrated in Table~\ref{appendix:tab_correlationperspective}. The results indicate that there is a lack of correlation between $(\theta_{T}\odot m)_{T}$ and $\theta_0$, but there is a correlation between $(\theta_{T}\odot m)_{T}$ and $\theta_T$. It further strengthens the conclusion that it is not desirable to have the weight correlation between final-trained weights and weight initialization.

\begin{table}[h!]
    \centering
    \caption{Weight correlation analysis at $p=0.2$ between $(\theta_{T}\odot m)_{T}$ and $\theta_0$, and between $(\theta_{T}\odot m)_{T}$ and $\theta_T$ for VGG-11, ResNet-20, MobileNet-V2 using CIFAR-10/100 at learning rate 0.1 under sparsity ratio 50\% and the masks $m$ are generated by ADMM-based pruning algorithm.}
    \label{appendix:tab_correlationperspective}
    \resizebox{0.95\linewidth}{!}{
    \begin{tabular}{cccc}
        \toprule
        Model & Dataset &$R_p((\theta_{T}\odot m)_{T}, \theta_0)$ & $R_p((\theta_{T}\odot m)_{T}, \theta_T)$ \\
        ResNet-20 & CIFAR-100 & 20.36\% & 63.97\% \\
        MobileNet-V2 & CIFAR-100 & 20.11\% & 64.71\% \\
        VGG-11 & CIFAR-100 & 20.41\% & 49.32\% \\
        MobileNet-V2 & CIFAR-10 & 20.26\% & 49.36\% \\
        VGG-11 & CIFAR-10 & 20.21\% & 48.08\%  \\
        \bottomrule
    \end{tabular}
    }
\end{table}

\section{Comparison with ~\cite{frankle2019stabilizing}}
\label{appendix:secG}

The work \citet{frankle2019stabilizing} suggests applying mask $m$ to $\theta_k$ and then apply sparse training, where $\theta_k$ denotes the weights trained from $\theta_0$ for a small number of $k$ epochs. This technique is training from $\theta_{k}\odot m$, and is in between sparse training (training from $\theta_{0}\odot m$) and pruning \& fine-tuning (training from $\theta_{T}\odot m$). We evaluate the relative sparse training performance among $(\theta_0 \odot m)_T$ (``winning ticket''), $(\theta_T \odot m)_T$ (pruned\&fine-tuned) and $(\theta_k \odot m)_T$ (``rewind'') under a desirable learning rate. We set $T=150$, $k=10$ and the initial learning rate is 0.1. The same hyperparameters are adopted as introduced in Section~\ref{subsec:revisitLT}. We study the accuracy performance comparison on MobileNet-V2, ResNet-20 and VGG-11 on CIFAR-100. We use the masks generated from the ADMM-based pruning algorithm. Figure~\ref{appendixfig:fig8_rewind} illustrates the accuracy comparison results of MobileNet-V2, ResNet-20 and VGG-11 on CIFAR-100. We can observe the order in the accuracy performance: $(\theta_T \odot m)_T$ (pruned\&fine-tuned) on top, $(\theta_k \odot m)_T$ (``rewind'') in the middle, and $(\theta_0 \odot m)_T$ (``winning ticket'') the lowest. 
As they exhibit the same number of training epochs (please note that $m$ is generated later than $\theta_k$ or $\theta_T$), we suggest directly applying the mask $m$ to $\theta_T$ and perform fine-tuning, instead of applying to $\theta_k$.

\end{document}